%% file: main_ConWriter.tex
\documentclass[11pt]{article}

\usepackage[final]{acl}

\usepackage{times}
\usepackage{latexsym}

\usepackage[T1]{fontenc}

\usepackage[utf8]{inputenc}

\usepackage{microtype}

\usepackage{inconsolata}

\usepackage{graphicx}

\usepackage{bbm}
\usepackage{centernot}  

\usepackage{titletoc}

\usepackage{multirow}
\usepackage{makecell}
\usepackage{booktabs}

\usepackage{algorithm}
\usepackage{algorithmicx}
\usepackage{algpseudocode}

\usepackage[table]{xcolor}

\definecolor{mycolor_darkblue}{RGB}{166, 202, 236}
\definecolor{mycolor_darkgreen}{RGB}{118, 197, 99}
\definecolor{mycolor_darkred}{RGB}{255, 191, 128}
\definecolor{mycolor_darkpink}{RGB}{245, 169, 176}

\definecolor{mycolor_lightblue}{RGB}{231, 247, 253}
\definecolor{mycolor_lightgreen}{RGB}{239, 251, 240}
\definecolor{mycolor_lightred}{RGB}{253, 246, 241}
\definecolor{mycolor_lightpink}{RGB}{253, 245, 252}

\definecolor{mycolor_ref}{RGB}{220, 0, 120}
\definecolor{mycolor_1}{RGB}{0, 0, 170}
\definecolor{mycolor_2}{RGB}{140, 70, 0}
\definecolor{mycolor_3}{RGB}{241, 243, 250}

\usepackage{xurl}
\usepackage{hyperref}

\usepackage{xcolor}
\usepackage{listings}
\usepackage[most]{tcolorbox}
\usepackage{needspace}
\tcbuselibrary{listings,breakable,skins}

\lstdefinelanguage{udiff}{
  morecomment=[f][\color{red!70!black}]{-},
  morecomment=[f][\color{green!50!black}]{+},
  morecomment=[f][\color{blue!60!black}]{@@},
  morecomment=[f][\color{gray!70}]{---},
  morecomment=[f][\color{gray!70}]{+++},
}

\lstdefinestyle{promptdiff}{
  language=udiff,
  basicstyle=\ttfamily\scriptsize,
  columns=fullflexible,
  keepspaces=true,
  showstringspaces=false,
  breaklines=true,
  breakatwhitespace=false,
  tabsize=2,
}

\newtcblisting{PromptDiffBoxGray}[2][]{%
  enhanced,
  breakable,
  colback=black!2,
  colframe=black!35,
  boxrule=0.35pt,
  arc=1.2mm,
  left=1.2mm,right=1.2mm,top=0.8mm,bottom=0.8mm,
  fonttitle=\bfseries\footnotesize,
  title={#2},
  listing only,
  listing options={
  style=promptdiff,
  escapeinside={(*@}{@*)}
  },
  #1
}

\newtcblisting{PromptDiffBoxBlue}[2][]{%
  enhanced,
  breakable,
  colback=mycolor_lightblue,
  colframe=mycolor_darkblue,
  boxrule=0.35pt,
  arc=1.2mm,
  left=1.2mm,right=1.2mm,top=0.8mm,bottom=0.8mm,
  fonttitle=\bfseries\footnotesize,
  title={#2},
  listing only,
  listing options={
  style=promptdiff,
  escapeinside={(*@}{@*)}
  },
  #1
}

\newtcblisting{PromptDiffBoxGreen}[2][]{%
  enhanced,
  breakable,
  colback=mycolor_lightgreen,
  colframe=mycolor_darkgreen,
  boxrule=0.35pt,
  arc=1.2mm,
  left=1.2mm,right=1.2mm,top=0.8mm,bottom=0.8mm,
  fonttitle=\bfseries\footnotesize,
  title={#2},
  listing only,
  listing options={
  style=promptdiff,
  escapeinside={(*@}{@*)}
  },
  #1
}

\newtcblisting{PromptDiffBoxRed}[2][]{%
  enhanced,
  breakable,
  colback=mycolor_lightred,
  colframe=mycolor_darkred,
  boxrule=0.35pt,
  arc=1.2mm,
  left=1.2mm,right=1.2mm,top=0.8mm,bottom=0.8mm,
  fonttitle=\bfseries\footnotesize,
  title={#2},
  listing only,
  listing options={
  style=promptdiff,
  escapeinside={(*@}{@*)}
  },
  #1
}

\newtcblisting{PromptDiffBoxPink}[2][]{%
  enhanced,
  breakable,
  colback=mycolor_lightpink,
  colframe=mycolor_darkpink,
  boxrule=0.35pt,
  arc=1.2mm,
  left=1.2mm,right=1.2mm,top=0.8mm,bottom=0.8mm,
  fonttitle=\bfseries\footnotesize,
  title={#2},
  listing only,
  listing options={
  style=promptdiff,
  escapeinside={(*@}{@*)}
  },
  #1
}

\title{ConWriter: Transition-Constrained Stateful Long-Form Story Generation with Lightweight Neuro-Symbolic Consistency Control}

\author{
  Jindong Li$^{1,a}$, 
  Yang Yang$^{1}$, 
  Zihao Liu$^{1}$, 
  Yutao Yue$^{1}$, 
  Menglin Yang$^{1,b}$\thanks{Corresponding author.} \\
  $^{1}$The Hong Kong University of Science and Technology (Guangzhou)  \\
  {$^{a}$jli839@connect.hkust-gz.edu.cn \quad $^{b}$menglinyang@hkust-gz.edu.cn}
}

\begin{document}
\maketitle

\input{sec-0_Abstract}

\input{sec-1_Introduction}


\input{sec-3_Method}

\input{sec-4_Experiment}

\input{sec-5_Conclusion}





\input{sec-6_Limitation-and-Ethical-Consideration}

\bibliography{main_ConWriter}

\clearpage
\appendix

\input{sec-7_Appendix}

\end{document}

%% file: sec-0_Abstract.tex
\begin{abstract}
Long-form story generation requires models to preserve narrative consistency across extended contexts, yet existing prompting-based methods often accumulate temporal, factual, character, commonsense, and stylistic errors as the story grows. We propose \textbf{\textsc{ConWriter}}, a training-free framework for consistency-aware long-form story generation. \textbf{\textsc{ConWriter}} writes stories incrementally at the scene level, guided by static story requirements, dynamic narrative memory, symbolic state reasoning, and uncertainty-aware risk signals. Rather than treating long-story generation as a single free-form decoding process, \textbf{\textsc{ConWriter}} maintains evolving story states, checks whether new scenes satisfy required narrative transitions, and uses uncertainty-aware risk signals to prioritize validation and localized repair. This enables consistency control during generation, before local errors propagate into later scenes. We evaluate \textbf{\textsc{ConWriter}} on ConStory-Bench across four long-story tasks, three target lengths, and multiple base LLMs following the official evaluation protocol. Across models and story lengths, \textbf{\textsc{ConWriter}} consistently matches or improves upon direct generation and outperforms the recent training-free baseline DOME in narrative consistency. These results demonstrate the effectiveness of lightweight neuro-symbolic consistency control for training-free long-form story generation. Code is available on \href{https://github.com/jindongli-Ai/ConWriter}{GitHub}.
\end{abstract}

%% file: sec-1_Introduction.tex
\begin{figure}[t]
    \centering
    \includegraphics[width=0.99\linewidth]{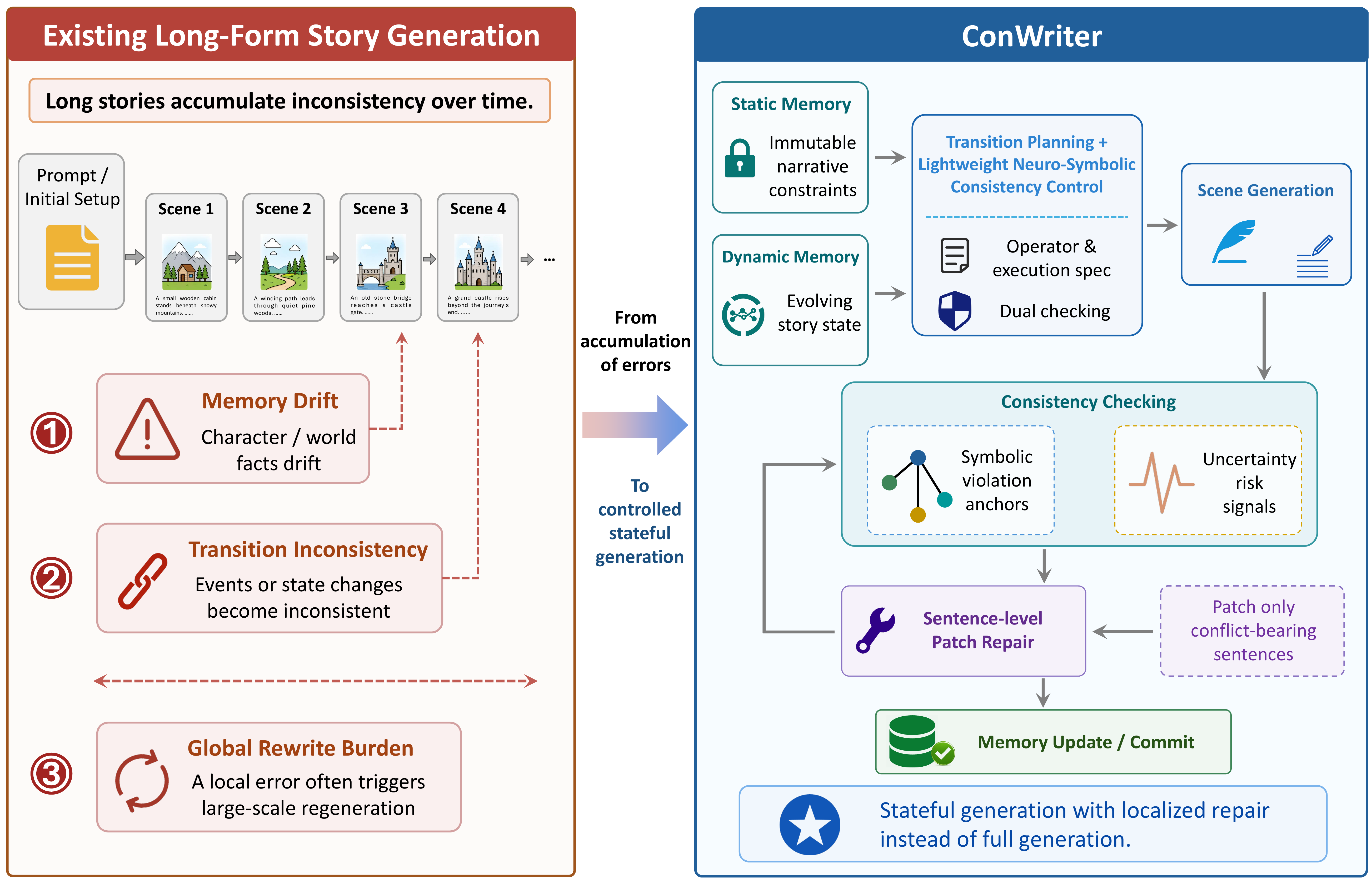}
    \caption{Challenges and motivation.}
    \label{fig:fig_1}
\end{figure}

\section{Introduction}
\label{sec:introduction}

Large language models (LLMs) \cite{2025_OpenAI_GPT-5-System-Card} have demonstrated strong capabilities in open-ended text generation, including creative writing and story generation \cite{2023_ACL_DOC_DOC--Improving-Long-Story-Coherence-with-Detailed-Outline-Control, 2025_NAACL_DOME_Generating-Long-form-Story-using-Dynamic-Hierarchical-Outlining-with-Memory-Enhancement}. However, long-form story generation remains a challenging setting: a model must maintain coherent characters, events, timelines, world facts, commonsense relations, and stylistic commitments across thousands of words. Unlike short-form generation, where local fluency is often sufficient, long stories require persistent global constraints and reliable state tracking. Recent benchmarks such as ConStory-Bench~\cite{2026_arXiv_ConStory-Bench-dataset_Lost-in-Stories--Consistency-Bugs-in-Long-Story-Generation-by-LLMs} show that even advanced LLMs still suffer from substantial \textit{consistency bugs} in extended narratives.

These consistency bugs become more severe as a story grows. A model may forget established facts, shift character attributes or motivations, violate temporal order, introduce implausible causal transitions, or drift from the intended style. ConStory-Bench~\cite{2026_arXiv_ConStory-Bench-dataset_Lost-in-Stories--Consistency-Bugs-in-Long-Story-Generation-by-LLMs} categorizes these failures into timeline, characterization, basic facts, commonsense, and style errors. Since local inconsistencies can propagate into later scenes, long-story generation requires mechanisms that track evolving narrative states and control consistency during generation.

Existing approaches provide only partial solutions. Outline- and plan-based methods introduce high-level structure for long stories~\cite{2023_ACL_DOC_DOC--Improving-Long-Story-Coherence-with-Detailed-Outline-Control, 2024_PAKDD_LongStory_LongStory--Coherent-Complete-and-Length-Controlled-Long-Story-Generation}, while memory-enhanced methods further organize long-form generation with dynamic outlines or persistent context~\cite{2025_NAACL_DOME_Generating-Long-form-Story-using-Dynamic-Hierarchical-Outlining-with-Memory-Enhancement, 2026_AAAI_Octopus_Octopus--Entropy-Controlled-Science-Fiction-Literature-Generation-with-Persistent-Memory-Context-Binding}. Other work studies constrained or reflection-driven long-form generation~\cite{2025_ACL-Findings_CogWriter_A-Cognitive-Writing-Perspective-for-Constrained-Long-Form-Text-Generation, 2025_arXiv_SuperWriter_SuperWriter--Reflection-Driven-Long-Form-Generation-with-Large-Language-Models}, and recent multi-agent systems decompose story writing into specialized agents~\cite{2025_CIKM_StoryWriter_StoryWriter--A-Multi-Agent-Framework-for-Long-Story-Generation}. However, planning, memory, or reflection alone does not explicitly verify whether each generated scene realizes a valid transition from the current narrative state. Long-story generation therefore still lacks a unified training-free framework that combines structured memory tracking, symbolic state validation, risk-aware checking, and targeted local repair during generation (see Figure~\ref{fig:fig_1}).

To address this problem, we propose \textbf{\textsc{ConWriter}}, a training-free framework for consistency-aware long story generation. The central idea is to treat long-story writing as a \textit{constraint-guided incremental generation process}. Instead of generating the whole story in a single pass, \textsc{ConWriter} writes the story scene by scene, maintaining static story requirements and dynamic narrative states throughout generation. At each step, it derives symbolic transition constraints, generates a scene under the current memory and constraints, and verifies whether the generated scene realizes a valid narrative transition before proceeding.

\textsc{ConWriter} integrates four lightweight control mechanisms. First, dual-memory modeling separates immutable story commitments from evolving events, entity states, relations, timelines, and pending constraints. Second, symbolic state-transition reasoning represents each scene as a checkable transition over narrative states. Third, dual consistency assurance combines structured validation with uncertainty-aware risk monitoring, allowing both explicit violations and weakly grounded segments to be inspected. Fourth, sentence-level patch repair revises only conflict-bearing sentences before accepted scenes are committed into dynamic memory. In this way, consistency control is integrated into the writing process rather than applied only after the full story has already been produced.

We evaluate \textbf{\textsc{ConWriter}} on ConStory-Bench~\cite{2026_arXiv_ConStory-Bench-dataset_Lost-in-Stories--Consistency-Bugs-in-Long-Story-Generation-by-LLMs}, covering four long-story tasks: continuation, generation, expansion, and completion. Due to the high cost of long-form generation and LLM-based evaluation, we use the first five cases from each task and test three target lengths: 3K, 6K, and 12K words. Experiments are conducted across three base LLM families, including Qwen3.5-Plus, DeepSeek-V4-Flash, and GPT-5.4-nano, following the official ConStory-Bench evaluation protocol. This setting allows us to study whether generation-time memory tracking, symbolic validation, uncertainty-aware checking, and local repair improve consistency across models and story lengths.
Our main contributions are summarized as follows:
\begin{itemize}
    \item We formulate long-story generation as a \textit{constraint-guided incremental writing problem}, where consistency is controlled during scene-level generation rather than only evaluated after generation.

    \item We propose \textbf{\textsc{ConWriter}}, a training-free framework with \textit{dual-memory stateful narrative modeling}, separating immutable static memory from evolving dynamic memory.

    \item We introduce \textit{lightweight neuro-symbolic consistency control with dual checking}, representing each scene as a state transition and validating it with structured checks and uncertainty-aware risk signals.

    \item We develop \textit{dual-guided sentence-level patch repair}, which uses symbolic violation anchors and uncertainty-aware risk signals to patch conflict-bearing sentences rather than regenerate entire scenes.

    \item We conduct extensive experiments on ConStory-Bench across four tasks, three target lengths, and multiple base LLMs, showing that \textbf{\textsc{ConWriter}} consistently matches or improves upon direct generation and outperforms the recent training-free baseline DOME in narrative consistency.
\end{itemize}

%% file: sec-3_Method.tex

\begin{figure*}
    \centering
    \includegraphics[width=0.99\linewidth]{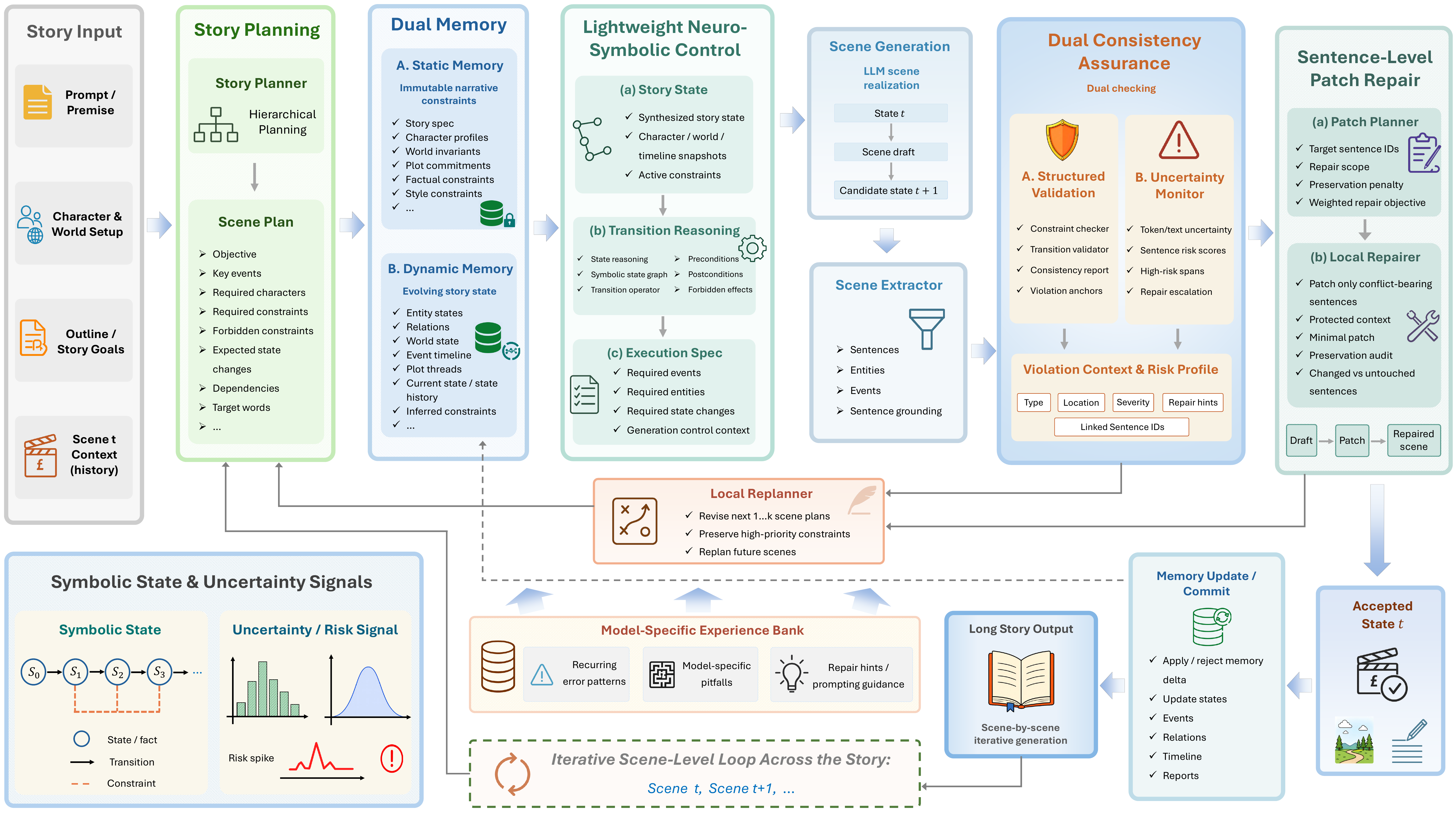}
    \caption{The overview of \textbf{\textsc{ConWriter}}.}
    \label{fig:framework}
\end{figure*}

\section{Methodology}
\label{sec:method}


\subsection{Problem Formulation}

Given a story specification $x$, long-story generation aims to produce a coherent narrative
$Y=\{y_1,\ldots,y_T\}$, where each $y_t$ denotes a scene. Unlike short-form generation, each scene should satisfy not only local fluency but also long-range consistency constraints over characters, events, temporal order, world facts, commonsense relations, and style. We formulate long-story generation as an incremental state-transition process:
\begin{equation}
    y_t \sim G_{\theta}
    \left(
    y_t \mid x, s_t, \mathcal{M}_t, \mathcal{C}_t
    \right),
\end{equation}
where  \(G_\theta\) denotes the base LLM, $s_t$ is the current scene specification, $\mathcal{M}_t$ is the narrative memory before scene $t$, and $\mathcal{C}_t$ denotes consistency constraints active at scene $t$.

The generated scene is accepted only if it induces a valid transition from the current story state to the next story state:
\begin{equation}
    \begin{gathered}
        \mathcal{M}_{t+1}
        =
        \operatorname{Update}(\mathcal{M}_{t}, y_t),
        \\
        \operatorname{Valid}
        \left(
        \mathcal{M}_{t}, y_t, \mathcal{M}_{t+1}
        \right)=1.
    \end{gathered}
\end{equation}
Thus, \textsc{ConWriter} treats long-story generation as constrained incremental writing rather than one-shot free-form decoding.


\subsection{Overview of \textsc{ConWriter}}

Figure~\ref{fig:framework} illustrates the overall framework of \textsc{ConWriter}. 
Rather than generating a long story in a single pass, \textsc{ConWriter} wraps a base LLM with generation-time consistency control. 
At each step, it grounds the current scene in static and dynamic memory, derives symbolic transition constraints, and asks the base LLM to generate a draft scene under these constraints and model-specific experience guidance. 
The draft is then examined by dual consistency assurance, where structured validation checks explicit transition violations and uncertainty-aware risk monitoring identifies unstable segments for stronger inspection. The uncertainty branch serves as an auxiliary risk signal and does not introduce a separately trained neural consistency verifier.
When necessary, \textsc{ConWriter} applies localized sentence-level repair and commits the scene to dynamic memory only after it passes consistency checks. Furthermore, Algorithm~\ref{alg:conwriter} in Appendix~\ref{sec:Appendix_Additional-Methods} provides the entire procedure of \textbf{\textsc{ConWriter}}.

Formally, each step is decomposed into five operations:
\begin{equation}
\begin{gathered}
    \mathcal{R}_t = \operatorname{Retrieve}(s_t,\mathcal{M}_t), \\
    o_t = \operatorname{Reason}(s_t,\mathcal{R}_t), \\
    \hat{y}_t = G_{\theta}(x,s_t,\mathcal{R}_t,o_t,\mathcal{E}_{t}), \\
    (v_t,r_t) = \operatorname{Assure}(\hat{y}_t,\mathcal{M}_t,o_t), \\
    y_t = \operatorname{AcceptOrRepair}(\hat{y}_t,v_t,r_t).
\end{gathered}
\end{equation}
Here, $\mathcal{R}_t$ is the bound memory context, $o_t$ is the symbolic transition operator, $\mathcal{E}_{t}$ denotes the scene-specific experience guidance selected from the model-specific experience bank $\mathcal{E}_{\theta}$, $v_t$ denotes explicit violation signals, and $r_t$ denotes uncertainty-aware risk signals. These operations define the core control loop of \textsc{ConWriter}: generation is allowed to proceed only after the current scene has been checked and, if necessary, repaired.


\subsection{Structured Narrative Memory}

\textsc{ConWriter} represents narrative memory as the combination of static and dynamic memory:
\begin{equation}
    \mathcal{M}_t =
    (
    \mathcal{M}^{s},
    \mathcal{M}^{d}_{t}
    ).
\end{equation}
Static memory $\mathcal{M}^{s}$ stores global commitments that should remain stable throughout the story:
\begin{equation}
    \mathcal{M}^{s}
    =
    \left\{
    \mathcal{P},
    \mathcal{W},
    \mathcal{A},
    \mathcal{S},
    \mathcal{I}
    \right\},
\end{equation}
where $\mathcal{P}$ is the premises, $\mathcal{W}$ denotes world rules, $\mathcal{A}$ denotes character attributes, $\mathcal{S}$ denotes style constraints, and $\mathcal{I}$ denotes task-specific instructions.

Dynamic memory $\mathcal{M}^{d}_{t}$ records the evolving story state before scene $t$:
\begin{equation}
    \mathcal{M}^{d}_{t}
    =
    \left(
    \mathcal{H}_{t},
    \mathcal{Z}_{t},
    \mathcal{R}^{ent}_{t},
    \mathcal{T}_{t},
    \mathcal{Q}_{t}
    \right),
\end{equation}
where $\mathcal{H}_{t}$ stores accumulated events, $\mathcal{Z}_{t}$ stores entity states, $\mathcal{R}^{ent}_{t}$ stores entity relations, $\mathcal{T}_{t}$ stores temporal information, and $\mathcal{Q}_{t}$ stores unresolved future constraints.

For scene $t$, \textbf{\textsc{ConWriter}} constructs a compact
scene-conditioned memory context:
\begin{equation}
    \mathcal{R}_t
    =
    \operatorname{Retrieve}(s_t,\mathcal{M}_t),
\end{equation}
where $\operatorname{Retrieve}(\cdot)$ selects the involved entities,
active constraints, recent events, unresolved plot threads, and
relevant static commitments for the current scene. In implementation,
the context size is controlled using deterministic filtering,
recency-based selection, and fixed top-$k$ caps over memory fields.


\subsection{Symbolic State-Transition Reasoning}

To make scene generation checkable, the model derives a symbolic transition operator for each scene:
\begin{equation}
    o_t =
    \left(
    \operatorname{Pre}_t,
    \operatorname{Post}_t,
    \operatorname{Forbid}_t
    \right).
\end{equation}
Here, $\operatorname{Pre}_t$ specifies conditions that should hold before writing the scene, $\operatorname{Post}_t$ specifies required state changes after the scene, and $\operatorname{Forbid}_t$ specifies invalid or contradictory changes.

The operator is feasible only if its preconditions are supported by the current memory:
\begin{equation}
    \operatorname{Feasible}(o_t,\mathcal{M}_t)
    =
    \mathbbm{1}
    \left[
    \forall c \in \operatorname{Pre}_t,
    \mathcal{M}_t \models c
    \right].
\end{equation}
Feasibility checking ensures that the framework does not ask the generator to execute a transition that contradicts the current story state.

The expected post-scene state is then represented as:
\begin{equation}
    \mathcal{M}^{\star}_{t+1}
    =
    \operatorname{Apply}
    \left(
    \mathcal{M}^{d}_{t},
    \operatorname{Post}_t
    \right).
\end{equation}
During generation, $o_t$ is converted into explicit writing constraints. During verification, the same operator becomes the basis for checking whether the generated scene realizes the intended transition.


\subsection{Experience-Guided Scene Generation}

The base LLM generates a draft scene under memory and transition constraints:
\begin{equation}
    \hat{y}_t = G_{\theta} \left(x, s_t, \mathcal{R}_t, o_t, \mathcal{E}_{t} \right),
\end{equation}
where $\mathcal{E}_{\theta}$ denotes the model-specific experience bank.
The bank stores recurring failure patterns from the base model $G_{\theta}$:
\begin{equation}
    \mathcal{E}_{\theta}
    =
    \left\{
    e_k=(a_k,c_k,h_k)
    \right\}_{k=1}^{K},
\end{equation}
where $a_k$ is an observed error pattern, $c_k$ is its applicable context, and $h_k$ is a corrective hint. For scene $t$, relevant experience items are selected by:
\begin{equation}
    \mathcal{E}_{t}
    =
    \left\{
    e_k \in \mathcal{E}_{\theta}
    \mid
    \operatorname{sim}(c_k,s_t,o_t,\mathcal{R}_t)>\tau_e
    \right\}.
\end{equation}
where $\tau_e$ is the experience-selection threshold. Here, $\operatorname{sim}(\cdot)$ denotes an implementation-level compatibility score over factors such as model identity, task type, generation stage, conflict type, frequency, confidence, and recency, rather than a learned embedding similarity. The selected items $\mathcal{E}_{t}$ are injected as lightweight guidance. They do not determine whether a scene is consistent; they only help the generator avoid known model-specific failure modes.


\subsection{Uncertainty-Aware Risk Monitoring}

Symbolic transition constraints provide explicit consistency checks, but they may miss uncertain or weakly grounded regions. 
To complement symbolic validation, \textsc{ConWriter} introduces an uncertainty-aware risk monitor as an auxiliary control signal. 
When model-side token log probabilities are available, the monitor estimates token-level uncertainty from the returned probability distribution. Otherwise, it falls back to a text-derived sentence-level uncertainty proxy based on lexical dispersion and punctuation density. In both cases, the resulting risk signal is used to prioritize validation and localized repair, rather than serving as a standalone consistency judgment. In the full \textsc{ConWriter} setting, the uncertainty layer remains enabled throughout generation; when token log probabilities are unavailable, the text-derived proxy is used instead.

Given a draft scene $\hat{y}_t=\{u_{t,1},\ldots,u_{t,n}\}$ composed of sentences, the monitor computes sentence-level risk score $\rho_{t,i}$ by the function $\operatorname{UncRisk}$:
\begin{equation}
    \rho_{t,i}
    =
    \operatorname{UncRisk}
    \left(
    u_{t,i},
    \mathcal{R}_t,
    o_t
    \right),
     i=1,\ldots,n.
\end{equation}
The scene-level risk $r_t$ is aggregated as:
\begin{equation}
    r_t
    =
    \alpha \cdot
    \frac{1}{n}\sum_{i=1}^{n}\rho_{t,i}
    +
    \beta \cdot
    \max_{i} \rho_{t,i}
    +
    \gamma \cdot
    \operatorname{Spike}(\rho_{t,1:n}),
\end{equation}
where the average term captures global instability, the maximum term captures localized high-risk sentences, and the spike term captures abrupt uncertainty changes, with $\alpha$, $\beta$, and $\gamma$ as weighting hyperparameters.

The risk signal $r_t$ is not used as a direct consistency judgment.
This distinction prevents the uncertainty monitor from being misinterpreted as a hard verifier. Instead, it controls the strength of later checking and repair:
\begin{equation}
    b_t
    =
    b_0
    +
    \eta \cdot \mathbbm{1}[r_t>\tau_r],
\end{equation}
where $b_t$ is the verification or repair budget allocated to scene $t$, and $\tau_r$ is the risk threshold used to allocate additional assurance budget to high-risk scenes. A high risk signal alone does not trigger repair.


\subsection{Consistency Verification}

After draft generation, \textsc{ConWriter} verifies whether the scene satisfies memory constraints and symbolic transition constraints. The violation score is defined as:
\begin{equation}
    V_t
    =
    V_{\mathrm{Pre}}+
    V_{\mathrm{Post}}+
    V_{\mathrm{Forbid}}+
    V_{\mathrm{Facet}},
\end{equation}
where each corresponds to different violation types.

Precondition violation checks whether the scene contradicts the required previous states:
\begin{equation}
    V_{\mathrm{Pre}}
    =
    \sum_{c\in \operatorname{Pre}_t}
    \mathbbm{1}
    \left[
    \hat{y}_t \not\models c
    \land
    \mathcal{M}_t \models c
    \right]. 
\end{equation}
Postcondition violation checks whether required state changes are realized:
\begin{equation}
    V_{\mathrm{Post}}
    =
    \sum_{c\in \operatorname{Post}_t}
    \mathbbm{1}
    \left[
    \hat{y}_t \not\models c
    \right].
\end{equation}
Forbidden-state violation checks whether the scene introduces invalid changes:
\begin{equation}
    V_{\mathrm{Forbid}}
    =
    \sum_{c\in \operatorname{Forbid}_t}
    \mathbbm{1}
    \left[
    \hat{y}_t \models c
    \right].
\end{equation}
Facet-level violation checks broader consistency categories:
\begin{equation}
    V_{\mathrm{Facet}}
    =
    \sum_{f\in \mathcal{F}}
    w_f
    \cdot
    \operatorname{Err}_{f}
    \left(
    \hat{y}_t,\mathcal{M}^{s},\mathcal{M}^{d}_{t}
    \right),
\end{equation}
where $\mathcal{F}$ includes timeline, characterization, basic facts, commonsense, style, etc.

The draft is accepted when no blocking consistency violation remains after assurance:
\begin{equation}
    \mathrm{Accept}(\hat{y}_t)=1,
    \quad \text{if} \enspace
    V_t = 0 .
\end{equation}

The resulting violation signals provide explicit anchors for localized repair. 
When any blocking violation remains, the draft is forwarded to the patch repair module together with its violated constraints and diagnostic evidence.


\subsection{Dual-Guided Sentence-level Patch Repair}

ConWriter repairs generated scenes at the sentence level, guided jointly by symbolic violation anchors and uncertainty-aware risk signals.

When blocking violations are detected, \textsc{ConWriter} constructs a repair specification, while uncertainty-aware risk signals are used to prioritize high-risk sentences during repair:

\begin{equation}
    \psi_t
    =
    \left(
    \mathcal{B}_t,
    \mathcal{D}_t,
    \mathcal{K}_t,
    \mathcal{N}_t
    \right),
\end{equation}
where $\mathcal{B}_t$ contains violated constraints, $\mathcal{D}_t$ contains diagnostic evidence, $\mathcal{K}_t$ contains content that must be preserved, and $\mathcal{N}_t$ contains forbidden changes.
The repair objective is:
\begin{equation}
    y_t
    =
    \arg\min_{y}
    \left[
    \operatorname{Dist}(y,\hat{y}_t)
    +
    \mu V(y)
    +
    \nu \operatorname{Risk}(y)
    \right],
\end{equation}
where $\operatorname{Dist}(y,\hat{y}_t)$ penalizes unnecessary rewriting, $V(y)$ measures remaining consistency violations, and $\operatorname{Risk}(y)$ measures residual uncertainty. This objective serves as a design-level repair criterion rather than an exact optimization over the full text space. In implementation, \textbf{\textsc{ConWriter}} approximates it through bounded sentence-level patch-target selection, localized LLM rewriting, and subsequent re-verification. The resulting bounded rewriting process is formulated as:
\begin{equation}
    \tilde{y}^{(j)}_t
    =
    G_{\theta}
    \left(
    \tilde{y}^{(j-1)}_t,
    \psi^{(j)}_t,
    \mathcal{R}_t,
    o_t,
    \mathcal{E}_{t}
    \right),
\end{equation}
where $\tilde{y}^{(0)}_t=\hat{y}_t$ and $j$ is the retry index. The bounded repair loop terminates when the repaired scene passes verification or the retry budget is exhausted. A repaired scene is accepted only when no blocking consistency violation remains:
\begin{equation}
    y_t
    =
    \tilde{y}^{(j)}_t,
    \quad
    \text{if}
    \enspace
    V\!\left(\tilde{y}^{(j)}_t\right)=0.
\end{equation}
If no candidate passes verification within the retry budget, the scene is rejected and does not update the dynamic memory.

\input{tabs/overall-performance}

\input{tabs/Ablation-Study}

\subsection{Dynamic Memory Update}

Only accepted scenes are allowed to update dynamic memory. Given the final scene $y_t$, \textsc{ConWriter} extracts state updates:
\begin{equation}
    \Delta \mathcal{M}^{d}_{t}
    =
    \operatorname{Extract}
    \left(
    y_t,
    \mathcal{M}^{s},
    \mathcal{M}^{d}_{t},
    o_t
    \right).
\end{equation}
The memory update is committed only when no blocking transition or constraint violation is detected in the candidate state:
\begin{equation}
    \operatorname{Commit}
    \left(
    \Delta \mathcal{M}^{d}_{t}
    \right)
    =
    \mathbbm{1}
    \left[
    V^{\mathrm{block}}_t = 0
    \right].
\end{equation}
Here, $V^{\mathrm{block}}_t$ denotes the blocking violations identified by the transition and constraint validators over the repaired scene and its candidate memory state.
Then the dynamic memory is updated as:
\begin{equation}
    \mathcal{M}^{d}_{t+1}
    =
    \mathcal{M}^{d}_{t}
    \oplus
    \Delta \mathcal{M}^{d}_{t},
\end{equation}
where $\oplus$ denotes conflict-aware memory merging. 

The final story is obtained by concatenating all accepted scenes:
\begin{equation}
    Y
    =
    \operatorname{Concat}
    \left(
    y_1,\ldots,y_T
    \right).
\end{equation}

Overall, these components form a closed-loop neuro-symbolic generation framework: each scene is planned, generated by the base LLM, checked against symbolic transition constraints, repaired when necessary, and committed only after passing consistency checks. 
By updating dynamic memory with only accepted scenes, \textsc{ConWriter} prevents local inconsistencies from being propagated into later story development. 
This design enables training-free, symbolic-state-guided consistency control during long-form generation, whose effectiveness is evaluated in the following experiments.

%% file: tabs/overall-performance.tex
\begin{table*}[!t]
\centering
\scalebox{0.56}{

\begin{tabular}{
>{\raggedright\arraybackslash}p{1.2cm}
>{\raggedright\arraybackslash}p{3.2cm}
>{\raggedright\arraybackslash}p{2.3cm} |
>{\centering\arraybackslash}p{1.3cm}
>{\centering\arraybackslash}p{1.5cm} |
>{\centering\arraybackslash}p{1.3cm}
>{\centering\arraybackslash}p{1.5cm} |
>{\centering\arraybackslash}p{1.3cm}
>{\centering\arraybackslash}p{1.5cm} |
>{\centering\arraybackslash}p{1.3cm}
>{\centering\arraybackslash}p{1.5cm} |
>{\centering\arraybackslash}p{1.4cm}
>{\centering\arraybackslash}p{2.6cm}
}

\toprule

\multirow{3}{*}{\textbf{\shortstack{Target\\Length}}} & \multirow{3}{*}{\textbf{\shortstack{Foundation\\Model}}} & \multirow{3}{*}{\textbf{Method}}
& \multicolumn{2}{c|}{\makecell[c]{\textbf{Task 1}\\ \textbf{Continuation}}} 
& \multicolumn{2}{c|}{\makecell[c]{\textbf{Task 2}\\ \textbf{Generation}}} 
& \multicolumn{2}{c|}{\makecell[c]{\textbf{Task 3}\\ \textbf{Expansion}}} 
& \multicolumn{2}{c|}{\makecell[c]{\textbf{Task 4}\\ \textbf{Completion}}}
& \multicolumn{2}{c}{\textbf{Overall}} \\

\cline{4-5} \cline{6-7} \cline{8-9} \cline{10-11} \cline{12-13}

& & 
& \makecell[c]{Avg. \\ Words} & \makecell[c]{Avg. \\ CED} ($\downarrow$)
& \makecell[c]{Avg. \\ Words} & \makecell[c]{Avg. \\ CED} ($\downarrow$)
& \makecell[c]{Avg. \\ Words} & \makecell[c]{Avg. \\ CED} ($\downarrow$)
& \makecell[c]{Avg. \\ Words} & \makecell[c]{Avg. \\ CED} ($\downarrow$)
& \makecell[c]{Avg. \\ Words} & \makecell[c]{Avg. \\ CED} ($\downarrow$) \\

\hline

\multirow{23}{*}{3K}
& Grok-4-Fast 
& Direct 
& 3183.8 & 0.6163 
& 3178.4 & 0.6303 
& 3140.8 & 1.8930 
& 3131.2 & 2.4257 
& 3158.55 & 1.3913 
\\
\cline{2-13}

& \multirow{1}{*}{Qwen3-32B} 
& Direct 
& 3153.2 & 0.6109 
& 3193.2 & 1.2771 
& 3145.2 & 0.0 
& 3178.8 & 0.0 
& 3167.60 & 0.4720 
\\
\cline{2-13}

& \multirow{1}{*}{DeepSeek-V3.2} 
& Direct 
& 3102.2 & 2.5691 
& 3083.6 & 0.6386 
& 3099.2 & 0.0 
& 3142.8 & 0.6623 
& 3106.95 & 0.9675 
\\
\cline{2-13}

& \multirow{1}{*}{GPT-4o-mini} 
& Direct 
& 3131.6 & 0.0 
& 3117.6 & 0.6441 
& 3105.4 & 0.0 
& 3157.0 & 1.2676
& 3127.90 & 0.4779 
\\
\cline{2-13}

& \multirow{1}{*}{GPT-5-mini} 
& Direct 
& 3163.6 & 0.0 
& 3131.8 & 0.6351 
& 3088.2 & 0.0 
& 3136.0 & 0.6274 
& 3129.90 & 0.3156 
\\
\cline{2-13}

& \multirow{3}{*}{GPT-5} 
& Direct 
& 3427.6 & 0.6371  
& 3188.8 & 0.0 
& 3334.0 & 0.6194 
& 3234.0 & 0.0 
& 3296.10 & 0.3141 
\\
\cline{3-13}
& & \cellcolor{mycolor_3}\textbf{\textsc{ConWriter}} 
& \cellcolor{mycolor_3}5217.8 & \cellcolor{mycolor_3}0.0 
& \cellcolor{mycolor_3}4728.4 & \cellcolor{mycolor_3}0.2733 
& \cellcolor{mycolor_3}6436.0 & \cellcolor{mycolor_3}0.0 
& \cellcolor{mycolor_3}4238.8 & \cellcolor{mycolor_3}0.0 
& \cellcolor{mycolor_3}5155.25 & \cellcolor{mycolor_3}0.0683 
\\
& & \cellcolor{mycolor_3}$\Delta_{CED}$ 
& \cellcolor{mycolor_3}- & \cellcolor{mycolor_3}\textcolor{mycolor_1}{-0.6371} 
& \cellcolor{mycolor_3}- & \cellcolor{mycolor_3}\textcolor{mycolor_2}{0.2733}
& \cellcolor{mycolor_3}- & \cellcolor{mycolor_3}\textcolor{mycolor_1}{-0.6194} 
& \cellcolor{mycolor_3}- & \cellcolor{mycolor_3}0.0 
& \cellcolor{mycolor_3}- & \cellcolor{mycolor_3}\textcolor{mycolor_1}{-0.2458 \textbf{\scriptsize (78.26\% $\downarrow$)}} 
\\
\cline{2-13}

& \multirow{5}{*}{Qwen3.5-Plus} 
& Direct 
& 3101.2 & 0.6629 
& 3028.2 & 0.0 
& 3065.4 & 0.0 
& 3564.2 & 0.0 
& 3189.75 & 0.1657 
\\
\cline{3-13}
& & DOME 
& 3463.4 & 1.1120
& 3322.4 & 1.2038
& 3549.6 & 0.0
& 3545.4 & 1.1024
& 3470.20 & 0.8546
\\
& & $\Delta_{CED}$ 
& - & \textcolor{mycolor_2}{0.4491}
& - & \textcolor{mycolor_2}{1.2038}
& - & 0.0
& - & \textcolor{mycolor_2}{1.1024}
& - & \textcolor{mycolor_2}{0.6889 \textbf{\scriptsize (415.75\% $\uparrow$)}}
\\
\cline{3-13}
& & \cellcolor{mycolor_3}\textbf{\textsc{ConWriter}} 
& \cellcolor{mycolor_3}5840.0 & \cellcolor{mycolor_3}0.1407 
& \cellcolor{mycolor_3}3471.4 & \cellcolor{mycolor_3}0.0 
& \cellcolor{mycolor_3}6311.4 & \cellcolor{mycolor_3}0.0 
& \cellcolor{mycolor_3}3820.8 & \cellcolor{mycolor_3}0.0 
& \cellcolor{mycolor_3}4860.90 & \cellcolor{mycolor_3}0.0352 
\\
& & \cellcolor{mycolor_3}$\Delta_{CED}$ 
& \cellcolor{mycolor_3}- & \cellcolor{mycolor_3}\textcolor{mycolor_1}{-0.5222} 
& \cellcolor{mycolor_3}- & \cellcolor{mycolor_3}0.0
& \cellcolor{mycolor_3}- & \cellcolor{mycolor_3}0.0
& \cellcolor{mycolor_3}- & \cellcolor{mycolor_3}0.0 
& \cellcolor{mycolor_3}- & \cellcolor{mycolor_3}\textcolor{mycolor_1}{-0.1305 \textbf{\scriptsize (78.76\% $\downarrow$)}} 
\\
\cline{2-13}

& \multirow{5}{*}{DeepSeek-V4-Flash} 
& Direct 
& 5030.2 & 2.3460 
& 4457.2 & 0.9877 
& 4326.4 & 0.4820 
& 4255.4 & 0.4492 
& 4517.30 & 1.0662 
\\
\cline{3-13}
& & DOME 
& 3315.6 & 1.2520
& 3369.0 & 0.0
& 4532.2 & 0.6050 
& 4085.4 & 0.9313
& 3825.55 & 0.6971
\\
& & $\Delta_{CED}$ 
& - & \textcolor{mycolor_1}{-1.0940}
& - & \textcolor{mycolor_1}{-0.9877} 
& - & \textcolor{mycolor_2}{0.1230}
& - & \textcolor{mycolor_2}{0.4821}
& - & \textcolor{mycolor_1}{-0.3691 \textbf{\scriptsize (34.62\% $\downarrow$)}}
\\
\cline{3-13}
& & \cellcolor{mycolor_3}\textbf{\textsc{ConWriter}} 
& \cellcolor{mycolor_3}3481.4 & \cellcolor{mycolor_3}0.5340 
& \cellcolor{mycolor_3}3245.0 & \cellcolor{mycolor_3}0.0 
& \cellcolor{mycolor_3}3662.6 & \cellcolor{mycolor_3}0.0 
& \cellcolor{mycolor_3}4181.0 & \cellcolor{mycolor_3}0.0 
& \cellcolor{mycolor_3}3642.50 & \cellcolor{mycolor_3}0.1335 
\\
& & \cellcolor{mycolor_3}$\Delta_{CED}$ 
& \cellcolor{mycolor_3}- & \cellcolor{mycolor_3}\textcolor{mycolor_1}{-1.8120} 
& \cellcolor{mycolor_3}- & \cellcolor{mycolor_3}\textcolor{mycolor_1}{-0.9877} 
& \cellcolor{mycolor_3}- & \cellcolor{mycolor_3}\textcolor{mycolor_1}{-0.4820} 
& \cellcolor{mycolor_3}- & \cellcolor{mycolor_3}\textcolor{mycolor_1}{-0.4492} 
& \cellcolor{mycolor_3}- & \cellcolor{mycolor_3}\textcolor{mycolor_1}{-0.9327 \textbf{\scriptsize (87.48\% $\downarrow$)}} 
\\
\cline{2-13}

& \multirow{5}{*}{GPT-5.4-nano} 
& Direct 
& 3120.2 & 0.0 
& 4605.2 & 0.0 
& 4268.0 & 0.0 
& 4592.2 & 0.0 
& 4146.40 & 0.0 
\\
\cline{3-13}
& & DOME 
& 3950.0 & 0.0
& 4263.2 & 0.0
& 4423.2 & 0.5165
& 4308.0 & 0.0
& 4236.10 & 0.1291
\\
& & $\Delta_{CED}$ 
& - & 0.0
& - & 0.0
& - & \textcolor{mycolor_2}{0.5165}
& - & 0.0
& - & \textcolor{mycolor_2}{0.1291 \textbf{\scriptsize (--\% $\uparrow$)}}
\\
\cline{3-13}
& & \cellcolor{mycolor_3}\textbf{\textsc{ConWriter}} 
& \cellcolor{mycolor_3}6451.8 & \cellcolor{mycolor_3}0.0 
& \cellcolor{mycolor_3}4974.2 & \cellcolor{mycolor_3}0.0 
& \cellcolor{mycolor_3}6382.2 & \cellcolor{mycolor_3}0.0 
& \cellcolor{mycolor_3}5196.4 & \cellcolor{mycolor_3}0.0 
& \cellcolor{mycolor_3}5751.15 & \cellcolor{mycolor_3}0.0 
\\
& & \cellcolor{mycolor_3}$\Delta_{CED}$ 
& \cellcolor{mycolor_3}- & \cellcolor{mycolor_3}0.0 
& \cellcolor{mycolor_3}- & \cellcolor{mycolor_3}0.0
& \cellcolor{mycolor_3}- & \cellcolor{mycolor_3}0.0
& \cellcolor{mycolor_3}- & \cellcolor{mycolor_3}0.0
& \cellcolor{mycolor_3}- & \cellcolor{mycolor_3}0.0 
\\

\hline
\hline


\multirow{21}{*}{6K}
& \multirow{1}{*}{Grok-4-Fast} 
& Direct
& 6232.4 & 0.0
& 6230.2 & 0.9546
& 6236.2 & 0.3257
& 6268.2 & 0.6350
& 6241.75 & 0.4788 
\\
\cline{2-13}

& \multirow{1}{*}{Qwen3-32B} 
& Direct 
& 6395.2 & 0.2783 
& 6997.8 & 0.9598 
& 6162.8 & 0.9833 
& 6218.4 & 2.5921 
& 6443.55 & 1.2034 
\\
\cline{2-13}

& \multirow{1}{*}{DeepSeek-V3.2} 
& Direct 
& 6062.0 & 0.3318 
& 6233.6 & 0.6191 
& 6220.0 & 0.3108 
& 6241.4 & 0.6403 
& 6189.25 & 0.4755 
\\
\cline{2-13}

& \multirow{1}{*}{GPT-4o-mini} 
& Direct 
& 6074.6 & 0.0 
& 6063.4 & 0.9868 
& 6120.6 & 0.3312 
& 6081.8 & 0.6596
& 6085.10 & 0.4944 
\\
\cline{2-13}

& \multirow{1}{*}{GPT-5-mini} 
& Direct 
& 6180.2 & 0.0 
& 6171.4 & 0.0 
& 6216.2 & 0.0 
& 6166.0 & 0.3292 
& 6183.45 & 0.0823 
\\
\cline{2-13}

& \multirow{1}{*}{GPT-5} 
& Direct 
& 6368.4 & 0.0 
& 6361.2 & 0.6258 
& 6329.6 & 0.0 
& 6394.0 & 0.0 
& 6363.30 & 0.1565 \\
\cline{2-13}

& \multirow{5}{*}{Qwen3.5-Plus} 
& Direct 
& 6268.6 & 0.0 
& 6077.4 & 0.0
& 7252.6 & 0.8251
& 7285.6 & 0.8183 
& 6721.05 & 0.4109
\\
\cline{3-13}
& & DOME 
& 6081.0 & 0.0
& 6393.6 & 0.6546
& 6473.8 & 0.6072
& 6207.2 & 0.6217
& 6288.90 & 0.4709 
\\
& & $\Delta_{CED}$ 
& - & 0.0
& - & \textcolor{mycolor_2}{0.6546}
& - & \textcolor{mycolor_1}{-0.2179}
& - & \textcolor{mycolor_1}{-0.1966}
& - & \textcolor{mycolor_2}{0.0600 \textbf{\scriptsize (14.60\% $\uparrow$)}}
\\
\cline{3-13}
& & \cellcolor{mycolor_3}\textbf{\textsc{ConWriter}} 
& \cellcolor{mycolor_3}6415.6 & \cellcolor{mycolor_3}0.0 
& \cellcolor{mycolor_3}6976.8 & \cellcolor{mycolor_3}0.0 
& \cellcolor{mycolor_3}9427.2 & \cellcolor{mycolor_3}0.2910 
& \cellcolor{mycolor_3}6829.4 & \cellcolor{mycolor_3}0.0 
& \cellcolor{mycolor_3}7412.25 & \cellcolor{mycolor_3}0.0728
\\
& & \cellcolor{mycolor_3}$\Delta_{CED}$ 
& \cellcolor{mycolor_3}- & \cellcolor{mycolor_3}0.0 
& \cellcolor{mycolor_3}- & \cellcolor{mycolor_3}0.0
& \cellcolor{mycolor_3}- & \cellcolor{mycolor_3}\textcolor{mycolor_1}{-0.5341}
& \cellcolor{mycolor_3}- & \cellcolor{mycolor_3}\textcolor{mycolor_1}{-0.8183} 
& \cellcolor{mycolor_3}- & \cellcolor{mycolor_3}\textcolor{mycolor_1}{-0.3381 \textbf{\scriptsize (82.28\% $\downarrow$)}} 
\\
\cline{2-13}

& \multirow{5}{*}{DeepSeek-V4-Flash} 
& Direct 
& 7981.6 & 1.1761 
& 8104.2 & 0.0 
& 8433.2 & 0.9619 
& 8576.0 & 0.4305 
& 8273.75 & 0.6421 
\\
\cline{3-13}
& & DOME 
& 7087.6 & 1.0973
& 7527.8 & 0.0
& 7154.8 & 0.0
& 6159.4 & 0.3316
& 6982.40 & 0.3572
\\
& & $\Delta_{CED}$ 
& - & \textcolor{mycolor_1}{-0.0788}
& - & 0.0
& - & \textcolor{mycolor_1}{-0.9619}
& - & \textcolor{mycolor_1}{-0.0989}
& - & \textcolor{mycolor_1}{-0.2849 \textbf{\scriptsize (44.37\% $\downarrow$)}}
\\
\cline{3-13}
& & \cellcolor{mycolor_3}\textbf{\textsc{ConWriter}} 
& \cellcolor{mycolor_3}6604.0 & \cellcolor{mycolor_3}0.3071 
& \cellcolor{mycolor_3}7209.4 & \cellcolor{mycolor_3}0.2841 
& \cellcolor{mycolor_3}6704.0 & \cellcolor{mycolor_3}0.3317 
& \cellcolor{mycolor_3}6764.2 & \cellcolor{mycolor_3}0.2643 
& \cellcolor{mycolor_3}6820.40 & \cellcolor{mycolor_3}0.2968 
\\
& & \cellcolor{mycolor_3}$\Delta_{CED}$ 
& \cellcolor{mycolor_3}- & \cellcolor{mycolor_3}\textcolor{mycolor_1}{-0.8690} 
& \cellcolor{mycolor_3}- & \cellcolor{mycolor_3}\textcolor{mycolor_2}{0.2841} 
& \cellcolor{mycolor_3}- & \cellcolor{mycolor_3}\textcolor{mycolor_1}{-0.6302} 
& \cellcolor{mycolor_3}- & \cellcolor{mycolor_3}\textcolor{mycolor_1}{-0.1662} 
& \cellcolor{mycolor_3}- & \cellcolor{mycolor_3}\textcolor{mycolor_1}{-0.3453 \textbf{\scriptsize (53.78\% $\downarrow$)}}  \\
\cline{2-13}

& \multirow{5}{*}{GPT-5.4-nano} 
& Direct 
& 6223.0 & 0.6538 
& 6448.0 & 0.3563 
& 6480.6 & 0.0 
& 6440.6 & 0.0 
& 6398.05 & 0.2525 \\
\cline{3-13}
& & DOME 
& 6678.0 & 0.0
& 6707.0 & 0.6088
& 6756.0 & 0.0
& 7353.4 & 0.0
& 6873.60 & 0.1522
\\
& & $\Delta_{CED}$ 
& - & \textcolor{mycolor_1}{-0.6538}
& - & \textcolor{mycolor_2}{0.2525}
& - & 0.0
& - & 0.0
& - & \textcolor{mycolor_1}{-0.1003 \textbf{\scriptsize (39.73\% $\downarrow$)}}
\\
\cline{3-13}
& & \cellcolor{mycolor_3}\textbf{\textsc{ConWriter}} 
& \cellcolor{mycolor_3}10605.6 & \cellcolor{mycolor_3}0.0883 
& \cellcolor{mycolor_3}8460.0 & \cellcolor{mycolor_3}0.0 
& \cellcolor{mycolor_3}8834.8 & \cellcolor{mycolor_3}0.1299 
& \cellcolor{mycolor_3}7810.4 & \cellcolor{mycolor_3}0.0 
& \cellcolor{mycolor_3}8927.70 & \cellcolor{mycolor_3}0.0546 
\\
& & \cellcolor{mycolor_3}$\Delta_{CED}$ 
& \cellcolor{mycolor_3}- & \cellcolor{mycolor_3}\textcolor{mycolor_1}{-0.5655}
& \cellcolor{mycolor_3}- & \cellcolor{mycolor_3}\textcolor{mycolor_1}{-0.3563}
& \cellcolor{mycolor_3}- & \cellcolor{mycolor_3}\textcolor{mycolor_2}{0.1299}
& \cellcolor{mycolor_3}- & \cellcolor{mycolor_3}0.0 
& \cellcolor{mycolor_3}- & \cellcolor{mycolor_3}\textcolor{mycolor_1}{-0.1979 \textbf{\scriptsize (78.38\% $\downarrow$)}} \\

\hline
\hline


\multirow{21}{*}{12K}
& \multirow{1}{*}{Grok-4-Fast} 
& Direct 
& 12169.6 & 0.3319
& 12194.8 & 0.4911
& 12265.4 & 0.4929
& 12146.0 & 0.3313
& 12193.95 & 0.4118 \\
\cline{2-13}

& \multirow{1}{*}{Qwen3-32B} 
& Direct 
& 12235.4 & 0.3282 
& 12395.2 & 0.6522
& 12247.4 & 0.1621 
& 12198.4 & 0.3318
& 12269.10 & 0.3686 \\
\cline{2-13}

& \multirow{1}{*}{DeepSeek-V3.2} 
& Direct 
& 12130.0 & 0.8260
& 12092.8 & 0.1652 
& 12297.2 & 0.3195 
& 12213.6 & 0.1666 
& 12183.40 & 0.3693 \\ 
\cline{2-13}

& \multirow{1}{*}{GPT-4o-mini} 
& Direct 
& 12143.8 & 0.0 
& 12105.6 & 0.6616 
& 12150.2 & 0.1663 
& 12138.2 & 0.1644
& 12134.45 & 0.2481 \\
\cline{2-13}

& \multirow{1}{*}{GPT-5-mini}
& Direct
& 12209.2 & 0.1611
& 12275.4 & 0.1590 
& 12151.6 & 0.0 
& 12321.0 & 0.1649 
& 12239.30 & 0.1213 \\
\cline{2-13}

& \multirow{1}{*}{GPT-5} 
& Direct 
& 12561.4 & 0.0 
& 12571.0 & 0.1585 
& 12449.6 & 0.3273
& 12366.2 & 0.3286 
& 12487.05 & 0.2036 \\
\cline{2-13}

& \multirow{5}{*}{Qwen3.5-Plus} 
& Direct 
& 12074.4 & 0.3327
& 12208.8 & 0.1639
& 12037.8 & 0.3328
& 12242.0 & 0.4916
& 12140.75 & 0.3303 
\\
\cline{3-13}
& & DOME 
& 12909.6 & 0.1639
& 12738.8 & 0.4833
& 12131.8 & 0.1708
& 12332.6 & 0.4907
& 12528.20 & 0.3272
\\
& & $\Delta_{CED}$ 
& - & \textcolor{mycolor_1}{-0.1688}
& - & \textcolor{mycolor_2}{0.3194}
& - & \textcolor{mycolor_1}{-0.1620}
& - & \textcolor{mycolor_1}{-0.0009}
& - & \textcolor{mycolor_1}{-0.0031 \textbf{\scriptsize (0.94\% $\downarrow$)}}
\\
\cline{3-13}
& & \cellcolor{mycolor_3}\textbf{\textsc{ConWriter}} 
& \cellcolor{mycolor_3}22015.2 & \cellcolor{mycolor_3}0.0526
& \cellcolor{mycolor_3}16252.6 & \cellcolor{mycolor_3}0.1454
& \cellcolor{mycolor_3}14961.0 & \cellcolor{mycolor_3}0.0
& \cellcolor{mycolor_3}12704.2 & \cellcolor{mycolor_3}0.4536
& \cellcolor{mycolor_3}16483.25 & \cellcolor{mycolor_3}0.1629
\\
& & \cellcolor{mycolor_3}$\Delta_{CED}$ 
& \cellcolor{mycolor_3}- & \cellcolor{mycolor_3}\textcolor{mycolor_1}{-0.2801}
& \cellcolor{mycolor_3}- & \cellcolor{mycolor_3}\textcolor{mycolor_1}{-0.0185}
& \cellcolor{mycolor_3}- & \cellcolor{mycolor_3}\textcolor{mycolor_1}{-0.3328}
& \cellcolor{mycolor_3}- & \cellcolor{mycolor_3}\textcolor{mycolor_1}{-0.0380}
& \cellcolor{mycolor_3}- &\cellcolor{mycolor_3}\textcolor{mycolor_1}{-0.1674 \textbf{\scriptsize (50.68\% $\downarrow$)}}
\\
\cline{2-13}

& \multirow{5}{*}{DeepSeek-V4-Flash} 
& Direct 
& 12100.0 & 0.4965 
& 12120.8 & 1.3187 
& 12083.0 & 0.1664 
& 12119.4 & 0.6591 
& 12105.80 & 0.6602 
\\
\cline{3-13}
& & DOME 
& 12143.6 & 0.8525
& 12929.2 & 0.4815
& 12452.8 & 0.4762
& 12455.0 & 1.1443
& 12495.15 & 0.7386
\\
& & $\Delta_{CED}$ 
& - & \textcolor{mycolor_2}{0.3560}
& - & \textcolor{mycolor_1}{-0.8372}
& - & \textcolor{mycolor_2}{0.3098}
& - & \textcolor{mycolor_2}{0.4852}
& - & \textcolor{mycolor_2}{0.0784 \textbf{\scriptsize (11.88\% $\uparrow$)}}
\\
\cline{3-13}
& & \cellcolor{mycolor_3}\textbf{\textsc{ConWriter}} 
& \cellcolor{mycolor_3}12479.8 & \cellcolor{mycolor_3}0.3262 
& \cellcolor{mycolor_3}12758.8 & \cellcolor{mycolor_3}0.3398 
& \cellcolor{mycolor_3}12225.0 & \cellcolor{mycolor_3}0.3303 
& \cellcolor{mycolor_3}12790.6 & \cellcolor{mycolor_3}0.3081 
& \cellcolor{mycolor_3}12563.55 & \cellcolor{mycolor_3}0.3261 
\\
& & \cellcolor{mycolor_3}$\Delta_{CED}$ 
& \cellcolor{mycolor_3}- & \cellcolor{mycolor_3}\textcolor{mycolor_1}{-0.1703} 
& \cellcolor{mycolor_3}- & \cellcolor{mycolor_3}\textcolor{mycolor_1}{-0.9789}
& \cellcolor{mycolor_3}- & \cellcolor{mycolor_3}\textcolor{mycolor_2}{0.1639} 
& \cellcolor{mycolor_3}- & \cellcolor{mycolor_3}\textcolor{mycolor_1}{-0.3510}
& \cellcolor{mycolor_3}- & \cellcolor{mycolor_3}\textcolor{mycolor_1}{-0.3341 \textbf{\scriptsize (50.61\% $\downarrow$)}} 
\\
\cline{2-13}

& \multirow{5}{*}{GPT-5.4-nano}
& Direct
& 12210.0 & 0.1598 
& 12250.4 & 0.0 
& 13164.6 & 0.0 
& 12344.4 & 0.1572 
& 12492.35 & 0.0793 
\\
\cline{3-13}
& & DOME 
& 12212.4 & 0.1676
& 12244.2 & 0.0
& 12179.4 & 0.0
& 12617.4 & 0.0
& 12313.35 & 0.0419
\\
& & $\Delta_{CED}$ 
& - & \textcolor{mycolor_2}{0.0078}
& - & 0.0
& - & 0.0
& - & \textcolor{mycolor_1}{-0.1572}
& - & \textcolor{mycolor_1}{-0.0374 \textbf{\scriptsize (47.16\% $\downarrow$)}} 
\\
\cline{3-13}
& & \cellcolor{mycolor_3}\textbf{\textsc{ConWriter}} 
& \cellcolor{mycolor_3}19334.8 & \cellcolor{mycolor_3}0.0 
& \cellcolor{mycolor_3}15315.0 & \cellcolor{mycolor_3}0.1296 
& \cellcolor{mycolor_3}16416.2 & \cellcolor{mycolor_3}0.0 
& \cellcolor{mycolor_3}14997.4 & \cellcolor{mycolor_3}0.0 
& \cellcolor{mycolor_3}16515.85 & \cellcolor{mycolor_3}0.0324 
\\
& & \cellcolor{mycolor_3}$\Delta_{CED}$ 
& \cellcolor{mycolor_3}- & \cellcolor{mycolor_3}\textcolor{mycolor_1}{-0.1598} 
& \cellcolor{mycolor_3}- & \cellcolor{mycolor_3}\textcolor{mycolor_2}{0.1296}
& \cellcolor{mycolor_3}- & \cellcolor{mycolor_3}0.0 
& \cellcolor{mycolor_3}- & \cellcolor{mycolor_3}\textcolor{mycolor_1}{-0.1572}
& \cellcolor{mycolor_3}- & \cellcolor{mycolor_3}\textcolor{mycolor_1}{-0.0469 \textbf{\scriptsize (59.14\% $\downarrow$)}} 
\\

\bottomrule

\end{tabular}
}
\caption{
Main results on ConStory-Bench~\cite{2026_arXiv_ConStory-Bench-dataset_Lost-in-Stories--Consistency-Bugs-in-Long-Story-Generation-by-LLMs} under different target story lengths.
All methods are evaluated under the \textit{forced-length setting }defined in Section~\ref{sec:Experiment_Experiment-Setup}, where each generated story must satisfy the specified target length.
Lower Avg. CED values indicate fewer consistency errors.
}
\label{tab:main_results}
\end{table*}

%% file: tabs/Ablation-Study.tex
\begin{table*}[t]
\centering
\scalebox{0.6}{

\begin{tabular}{
>{\raggedright\arraybackslash}p{5cm} |
>{\centering\arraybackslash}p{1.4cm}
>{\centering\arraybackslash}p{1.5cm} |
>{\centering\arraybackslash}p{1.4cm}
>{\centering\arraybackslash}p{1.5cm} |
>{\centering\arraybackslash}p{1.4cm}
>{\centering\arraybackslash}p{1.5cm} |
>{\centering\arraybackslash}p{1.4cm}
>{\centering\arraybackslash}p{1.5cm} |
>{\centering\arraybackslash}p{1.4cm}
>{\centering\arraybackslash}p{1.5cm}
}

\toprule
\multirow{3}{*}{\textbf{Variant}} 
& \multicolumn{2}{c|}{\makecell[c]{\textbf{Task 1}\\ \textbf{Continuation}}} 
& \multicolumn{2}{c|}{\makecell[c]{\textbf{Task 2}\\ \textbf{Generation}}} 
& \multicolumn{2}{c|}{\makecell[c]{\textbf{Task 3}\\ \textbf{Expansion}}} 
& \multicolumn{2}{c|}{\makecell[c]{\textbf{Task 4}\\ \textbf{Completion}}}
& \multicolumn{2}{c}{\textbf{Overall}} \\

\cline{2-11}

& \makecell[c]{Avg. \\ Words} & \makecell[c]{Avg. \\ CED} ($\downarrow$)
& \makecell[c]{Avg. \\ Words} & \makecell[c]{Avg. \\ CED} ($\downarrow$)
& \makecell[c]{Avg. \\ Words} & \makecell[c]{Avg. \\ CED} ($\downarrow$)
& \makecell[c]{Avg. \\ Words} & \makecell[c]{Avg. \\ CED} ($\downarrow$)
& \makecell[c]{Avg. \\ Words} & \makecell[c]{Avg. \\ CED} ($\downarrow$) \\

\hline
\cellcolor{mycolor_3}Full \textsc{ConWriter} 
& \cellcolor{mycolor_3}3481.4 & \cellcolor{mycolor_3}0.5340 
& \cellcolor{mycolor_3}3245.0 & \cellcolor{mycolor_3}0.0 
& \cellcolor{mycolor_3}3662.6 & \cellcolor{mycolor_3}0.0 
& \cellcolor{mycolor_3}4181.0 & \cellcolor{mycolor_3}0.0 
& \cellcolor{mycolor_3}3642.5 & \cellcolor{mycolor_3}0.1335 
\\

w/o Dynamic Memory 
& 8554.2 & 0.2938 
& 5409.0 & 0.9135 
& 6045.0 & 1.6309 
& 6942.6 & 0.1615 
& 6737.7 & 0.7499 
\\

w/o Structured Validation 
& 8395.8 & 0.6143 
& 6773.0 & 0.4481 
& 8885.6 & 0.1754 
& 6842.0 & 1.4995 
& 7724.1 & 0.6843 
\\

w/o Uncertainty Monitoring 
& 8839.6 & 1.0350 
& 8654.6 & 0.6471 
& 8057.0 & 0.3106 
& 4910.8 & 0.1925 
& 7615.5 & 0.5463 
\\

\hline
\end{tabular}
}
\caption{
Ablation results of \textsc{ConWriter} under the 3K \textit{forced-length setting} with DeepSeek-V4-Flash.
}
\label{tab:ablation}
\end{table*}

%% file: sec-4_Experiment.tex
\section{Experiment}
\label{sec:Experiment}

\begin{figure}[t]
    \centering
    \includegraphics[width=0.99\linewidth]{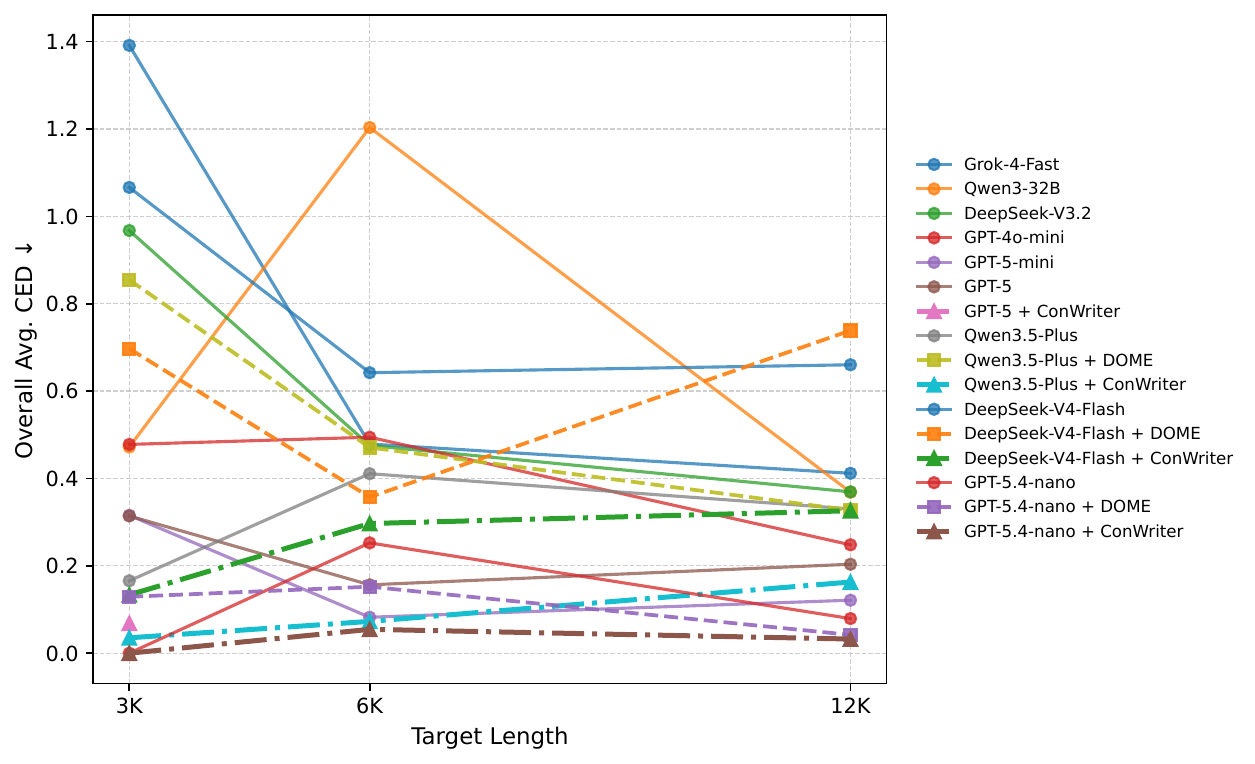}
    \caption{All models Avg. CED versus length.}
    \label{fig:All-models_CED-Length}
\end{figure}

\begin{figure}[t]
    \centering
    \includegraphics[width=0.99\linewidth]{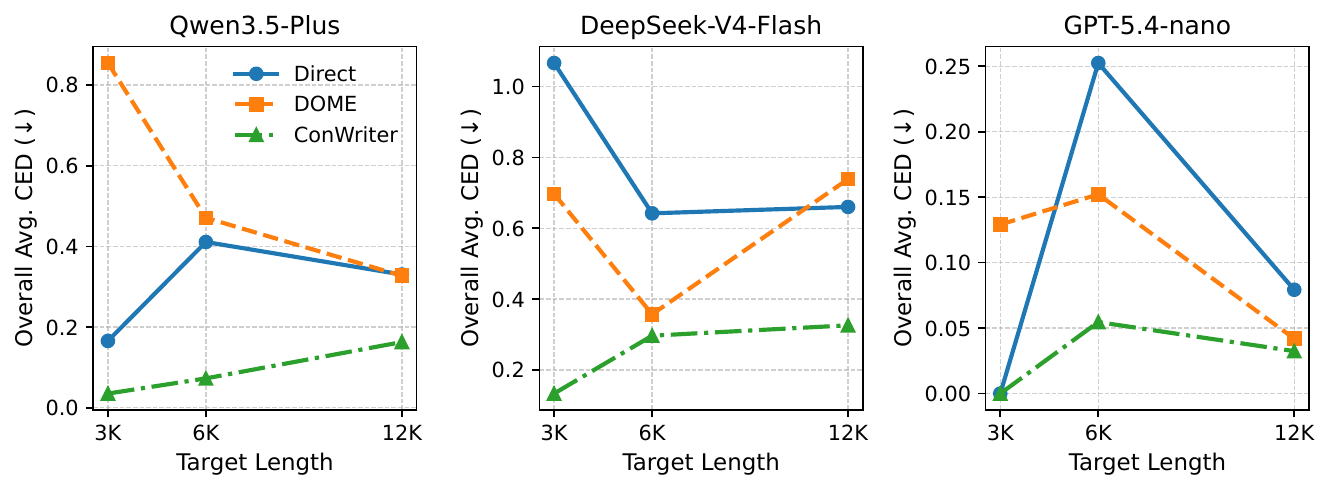}
    \caption{Overall Avg. CED across target lengths. \textbf{\textsc{ConWriter}} consistently yields lower CED across the three base LLM families.}
    \label{fig:avg-CED-vs-Length}
\end{figure}

\begin{figure*}[t]
    \centering
    \includegraphics[width=0.23\linewidth]{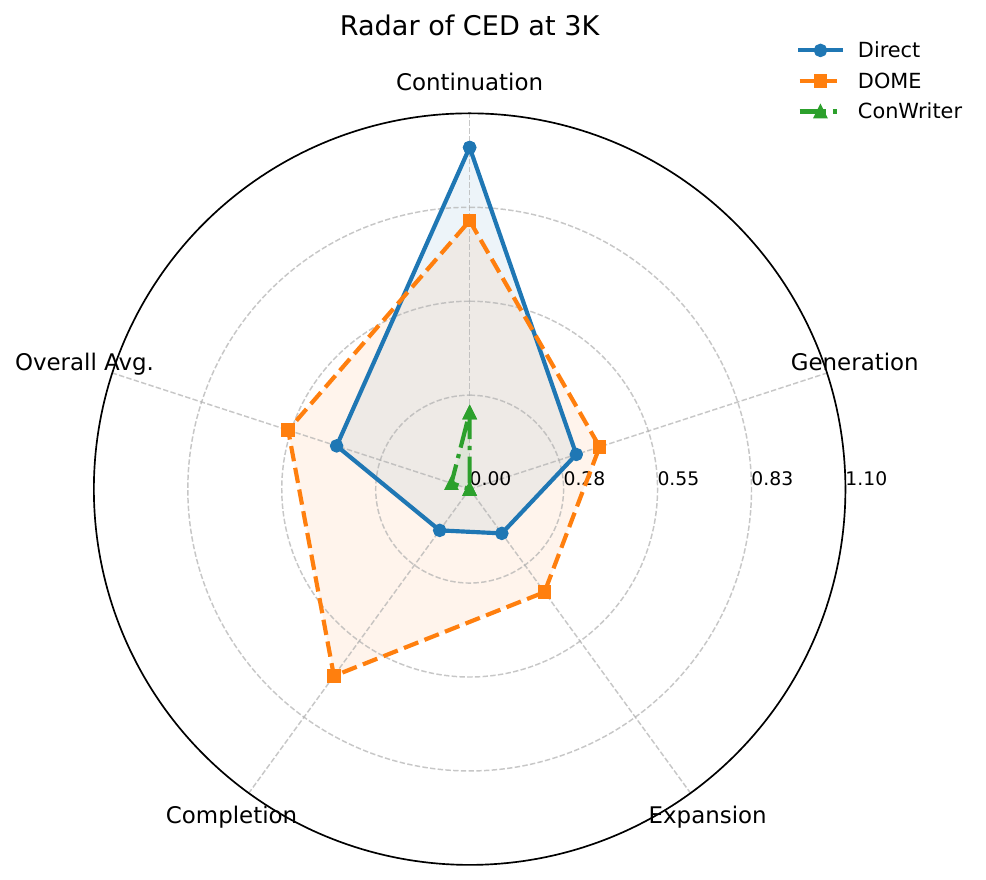}
    \includegraphics[width=0.23\linewidth]{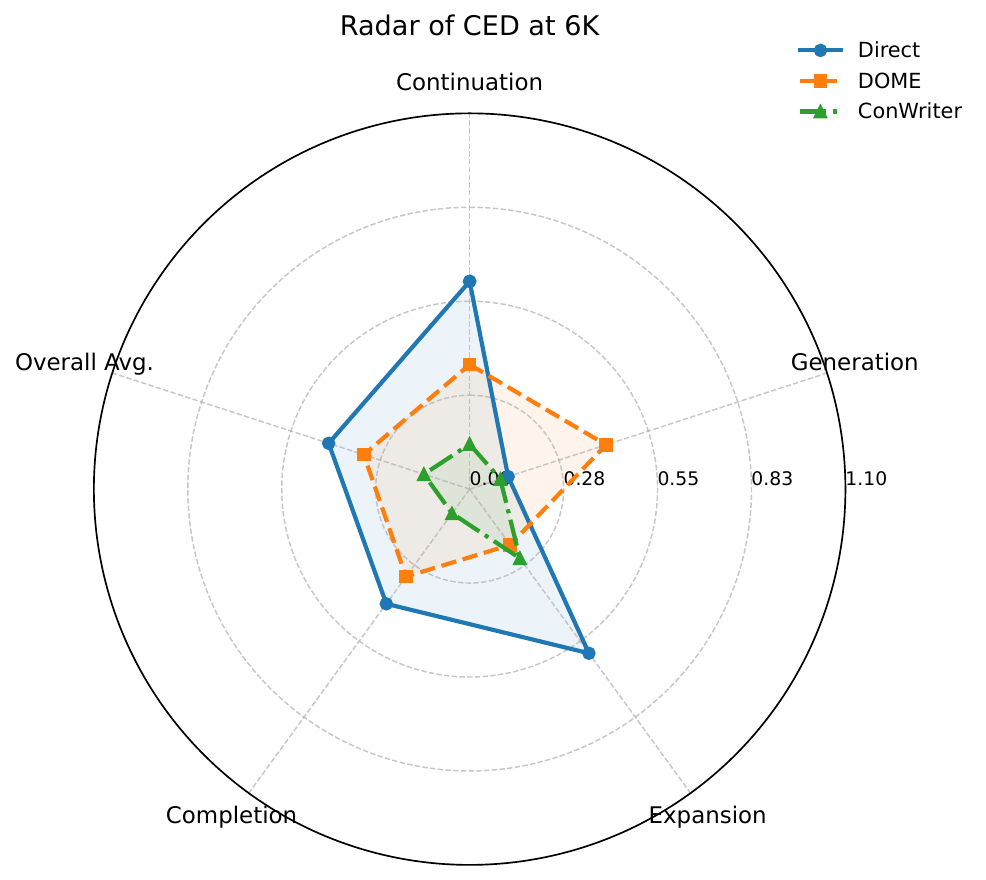}
    \includegraphics[width=0.23\linewidth]{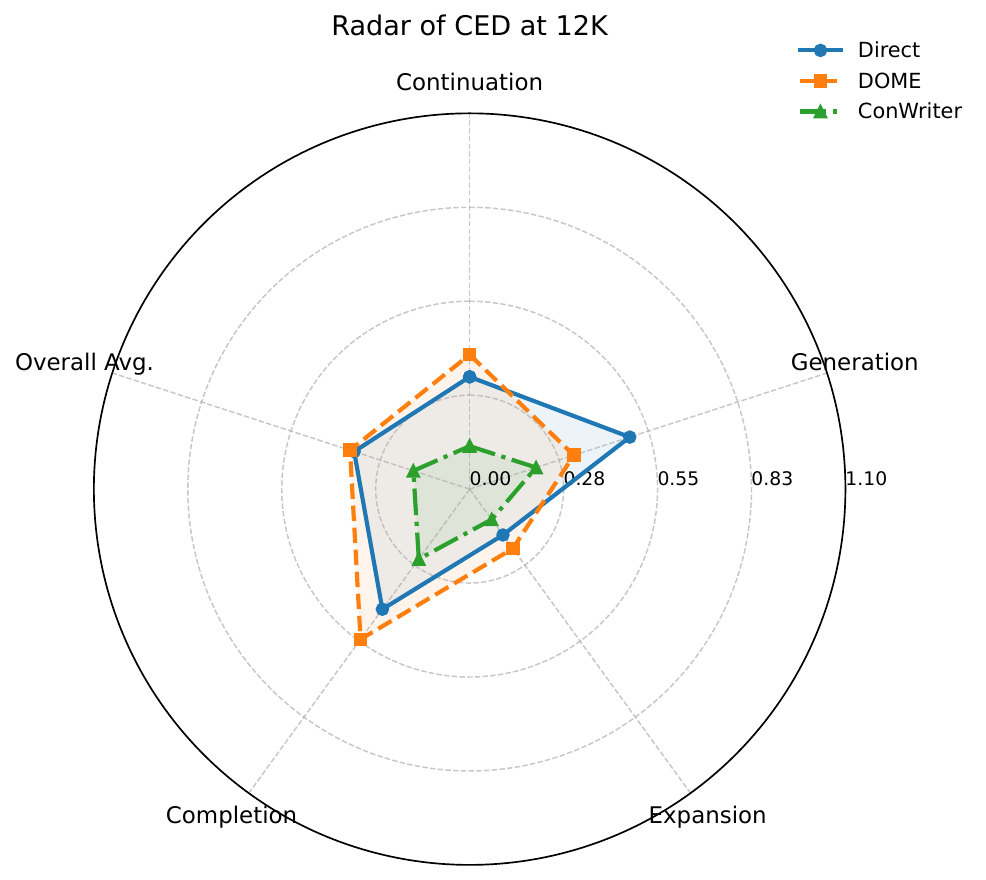}
    \includegraphics[width=0.24\linewidth]{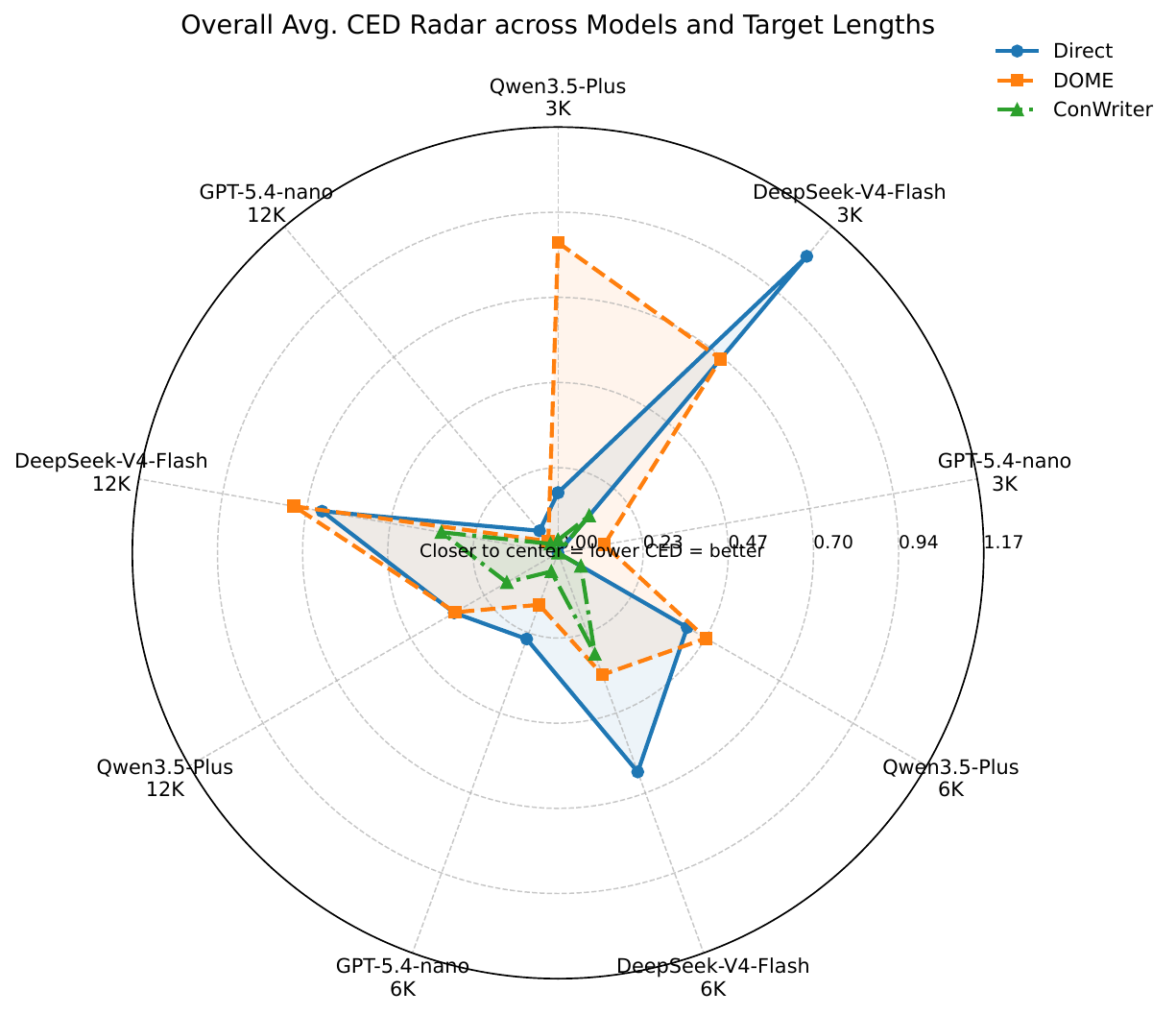}
    \caption{Radar views of CED across tasks, target lengths, and base LLM families. Lower values closer to the center indicate better consistency.}
    \label{fig:radar}
\end{figure*}

\begin{figure}[t]
    \centering
    \includegraphics[width=0.45\linewidth]{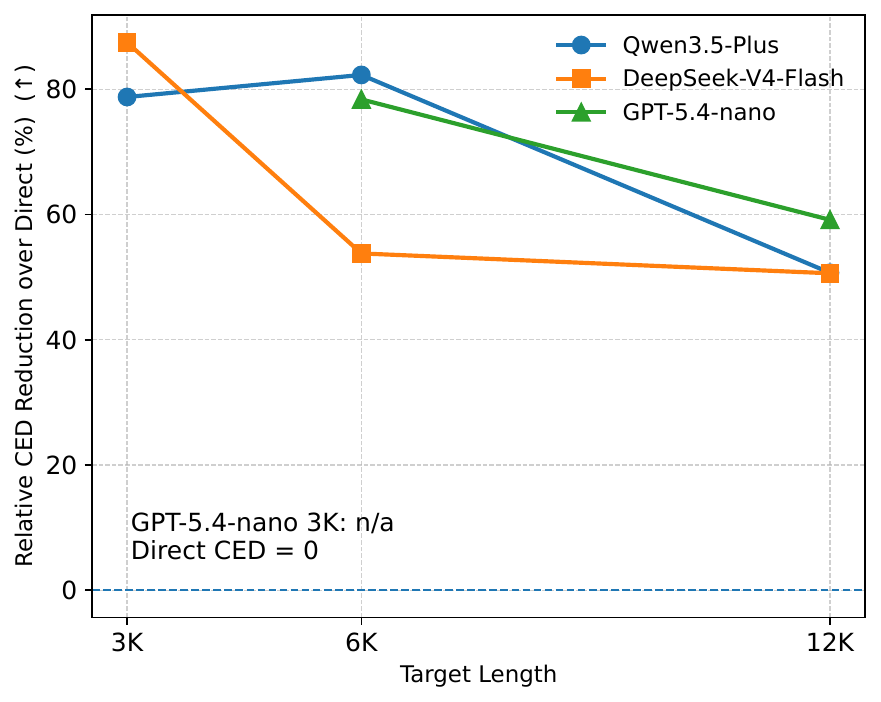}
    \hspace{0.3em}
    \includegraphics[width=0.45\linewidth]{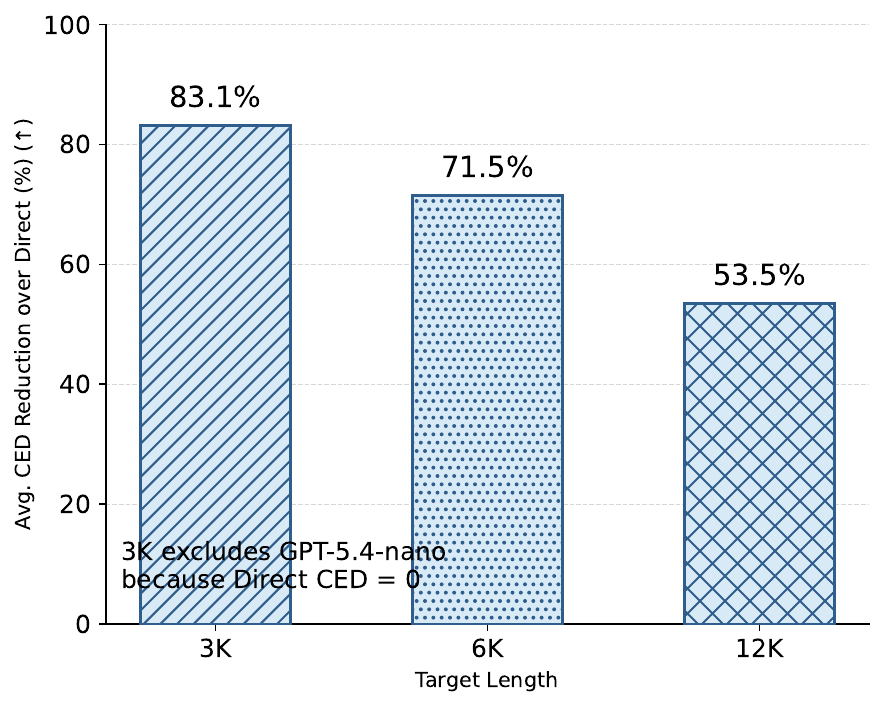}
    \caption{Relative CED reduction of \textbf{\textsc{ConWriter}} over Direct across target lengths.}
    \label{fig:relative-CED-reduction}
\end{figure}

\subsection{Experimental Setup}
\label{sec:Experiment_Experiment-Setup}

\paragraph{Dataset.}
We evaluate \textbf{\textsc{ConWriter}} on ConStory-Bench~\cite{2026_arXiv_ConStory-Bench-dataset_Lost-in-Stories--Consistency-Bugs-in-Long-Story-Generation-by-LLMs}, a benchmark designed for long-form story generation. Since long-form story generation and consistency evaluation are computationally expensive, prior work (e.g., DOC \cite{2023_ACL_DOC_DOC--Improving-Long-Story-Coherence-with-Detailed-Outline-Control}, DOME~\cite{2025_NAACL_DOME_Generating-Long-form-Story-using-Dynamic-Hierarchical-Outlining-with-Memory-Enhancement}) evaluates on 20 generated long stories. Following this controlled evaluation setting, we select the first five cases from each of the four tasks, resulting in 20 long-story evaluation cases in total. More details are provided in Appendix \ref{app:Experiment_Dataset}.

\paragraph{Target lengths.}
We evaluate each method with minimum target lengths of 3K, 6K, and 12K words. Unlike the original ConStory-Bench setting, we adopt a forced-length setting where each story is required to reach at least the specified minimum target length, while outputs are allowed to exceed it. This makes consistency control more challenging. More details are provided in Appendix~\ref{sec:Appendix_Experiment_Setup_Target-Length}.

\paragraph{Baselines.}
\textbf{\textit{Direct Generation.}}
The base LLM directly generates the full story from the input specification, including Grok-4-Fast \cite{2025_xAI_Grok-4-Fast}, Qwen3-32B \cite{2025_arXiv_Qwen3_Qwen3-Technical-Report}, DeepSeek-V3.2 \cite{2025_arXiv_DeepSeek-V3.2_DeepSeek-V3.2--Pushing-the-Frontier-of-Open-Large-Language-Models}, GPT-4o-mini \cite{2024_OpenAI_GPT-4o-mini}, GPT-5-mini \cite{2025_OpenAI_GPT-5-System-Card} and GPT-5 \cite{2025_OpenAI_GPT-5-System-Card}.
\textbf{\textit{Training-Free Baseline.}}
We use DOME~\cite{2025_NAACL_DOME_Generating-Long-form-Story-using-Dynamic-Hierarchical-Outlining-with-Memory-Enhancement} as a recent training-free baseline for long story generation.

\paragraph{Base models.}
We test three base LLM families: Qwen3.5-Plus \cite{2026_Web_Qwen3.5_Qwen3.5--Towards-Native-Multimodal-Agents, 2026_Alibaba-Cloud_Alibaba-Cloud-Model-Studio--Model-List}, DeepSeek-V4-Flash \cite{2026_arXiv_DeepSeek-V4_DeepSeek-V4--Towards-Highly-Efficient-Million-Token-Context-Intelligence}, and GPT-5.4-nano \cite{2026_OpenAI_GPT-5.4-mini-and-nano}. For each base model, we compare direct generation with the corresponding \textbf{\textsc{ConWriter}}-wrapped generation.

\paragraph{Evaluation metrics.}
We follow the ConStory-Bench~\cite{2026_arXiv_ConStory-Bench-dataset_Lost-in-Stories--Consistency-Bugs-in-Long-Story-Generation-by-LLMs} evaluation protocol and report the overall consistency error density and facet-level error densities. Following ConStory-Bench, we use o4-mini~\cite{2025_OpenAI-System-Card_o4-mini_OpenAI-o3-and-o4-mini-System-Card} as the evaluation model in ConStory-Checker to detect and categorize consistency errors. Lower values indicate better consistency.
More details are in Appendix~\ref{app:Experiment_Evaluation}.

\subsection{Main Results}

Table~\ref{tab:main_results} reports the full results under the forced-length setting. Overall, \textbf{\textsc{ConWriter}} consistently matches or reduces Overall Avg. CED across the three wrapped base LLM families and all target lengths. For Qwen3.5-Plus, it reduces Avg. CED by 78.76\%, 82.28\%, and 50.68\% under 3K, 6K, and 12K, respectively. For DeepSeek-V4-Flash, the reductions are 87.48\%, 53.78\%, and 50.61\%. For GPT-5.4-nano, direct generation already reaches 0.0 Avg. CED at 3K, but \textbf{\textsc{ConWriter}} preserves this ceiling score and further reduces Avg. CED by 78.38\% and 59.14\% at 6K and 12K.

The gains are larger when the base model exhibits clear consistency errors. DeepSeek-V4-Flash has relatively high direct-generation CED, and \textbf{\textsc{ConWriter}} substantially reduces it across all lengths. In contrast, GPT-5.4-nano is already highly consistent in several settings, leaving less room for improvement. This suggests that \textbf{\textsc{ConWriter}} acts as a generation-time consistency-control layer: it provides larger gains when narrative errors are frequent, while avoiding degradation when the base model is already strong.

Figures~\ref{fig:All-models_CED-Length}, \ref{fig:avg-CED-vs-Length}, and \ref{fig:radar} provide complementary visual evidence. They show that length-dependent inconsistency remains under forced-length generation, and that \textbf{\textsc{ConWriter}} stays more stable than Direct and DOME across model families and target lengths. Figure~\ref{fig:relative-CED-reduction} further shows that \textbf{\textsc{ConWriter}} achieves substantial relative reductions over Direct, although the average relative gain decreases as the target length increases, indicating that longer stories remain more challenging. Compared with DOME, which sometimes increases CED, \textbf{\textsc{ConWriter}} consistently lowers Overall Avg. CED, supporting the importance of generation-time memory tracking, transition validation, and localized repair.
More details are in Appendix~\ref{sec:Appendix_Experiment_Main-Results}.

\begin{figure}[t]
    \centering
    \includegraphics[width=0.99\linewidth]{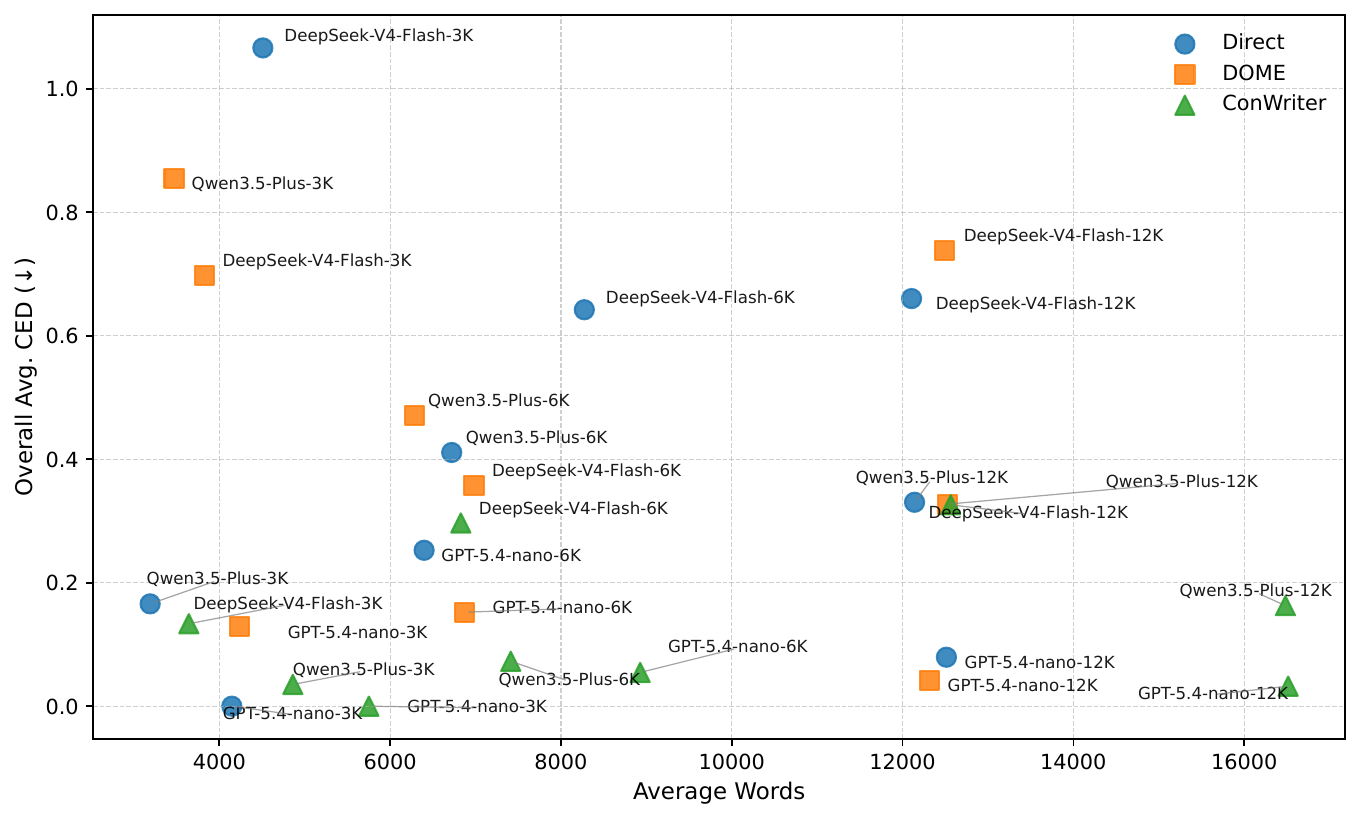}
    \caption{Average words versus overall Avg. CED. 
    }
    \label{fig:words_vs_ced_scatter_adjusted_labels}
\end{figure}

\subsection{Ablation Study}

Table~\ref{tab:ablation} shows that removing any major component degrades consistency. Dynamic memory is necessary for tracking evolving narrative states, structured validation makes scene transitions checkable, and uncertainty monitoring helps identify weakly grounded segments for stronger inspection or repair. The full \textbf{\textsc{ConWriter}} achieves the lowest Overall Avg. CED, confirming that the gains come from the combined control loop rather than a single isolated module. More details are provided in Appendix~\ref{sec:Appendix_Experiment_Ablation-Study}.

\subsection{Analysis and Discussion}

Figure~\ref{fig:words_vs_ced_scatter_adjusted_labels} compares average output length with Overall Avg. CED. \textbf{\textsc{ConWriter}} often generates comparable or longer stories while maintaining lower CED, showing that its gains are not caused by shorter outputs.
The results also reveal a base-model-dependent effect. \textbf{\textsc{ConWriter}} brings larger gains when direct generation has clear consistency errors, such as DeepSeek-V4-Flash, while the improvement is naturally smaller under ceiling-effect settings such as GPT-5.4-nano at 3K. This suggests that \textbf{\textsc{ConWriter}} acts as a robustness layer for generation-time consistency control.

\noindent\textbf{Discussion.}
Planning and memory are helpful only when reliably controlled: for weaker LLMs they provide useful structure, while for stronger LLMs a loose pipeline may introduce noise, as seen in some DOME settings. This motivates our training-free agent-style design, which targets strong contemporary LLMs and improves consistency through state tracking, transition validation, and local repair without model training. Although longer stories usually increase consistency risk, CED is not strictly monotonic due to model-specific behavior and length normalization. More details are in Appendix~\ref{sec:Appendix_Experiment_Discussion}.

%% file: sec-5_Conclusion.tex
\section{Conclusion}
\label{sec:conclusion}

We propose \textsc{ConWriter}, a training-free framework for consistency-aware long story generation. \textsc{ConWriter} formulates long-form writing as an incremental state-transition process and integrates structured memory, symbolic reasoning, uncertainty-aware risk monitoring, model-specific experience guidance, and local repair. Experiments on ConStory-Bench evaluate its effectiveness across four tasks, multiple story lengths, and different base LLMs.

%% file: sec-6_Limitation-and-Ethical-Consideration.tex
\section*{Limitations}

This work has several limitations. First, due to the high cost of long-form story generation and official consistency evaluation, we conduct experiments on a controlled subset of ConStory-Bench. This setting allows us to compare multiple target lengths and base LLM families under the same protocol, but future work can further scale the evaluation to more prompts and narrative domains. Second, \textbf{\textsc{ConWriter}} introduces additional inference overhead because it performs scene-level planning, validation, risk monitoring, and localized repair. This cost is a natural trade-off for generation-time consistency control. Third, the framework is most suitable for capable base LLMs that can follow long-form writing, correction, and length-control instructions. For some strong but less repair-compliant models, repeated feedback from the checking module may conflict with length control or targeted revision. Finally, since high-quality storytelling is the primary goal, we focus on contemporary strong LLMs rather than weak generators, where failures may reflect general generation limitations rather than long-story consistency control.

\section*{Ethical Considerations}

\textbf{\textsc{ConWriter}} is designed for consistency-aware long-form story generation and creative writing assistance. Like other LLM-based generation systems, it may inherit biases, unsafe assumptions, or hallucinated details from the base model. Its consistency-control mechanisms reduce narrative contradictions but do not guarantee factuality, fairness, or safety. Users should avoid applying the system to generate deceptive, harmful, or copyright-infringing content. Our experiments use the official ConStory-Bench protocol and evaluate generated stories only for research purposes.

%% file: sec-7_Appendix.tex
\startcontents[sections]
\printcontents[sections]{}{0}{\section*{Appendix Contents}\setcounter{tocdepth}{3}}

\input{sec-2_Related-Work}

\section{Additional Methods}
\label{sec:Appendix_Additional-Methods}

Algorithm~\ref{alg:conwriter} summarizes the inference procedure of \textbf{\textsc{ConWriter}}. Given a story input, the framework first constructs scene specifications and initializes static and dynamic memory. For each scene, it retrieves relevant memory, derives symbolic transition constraints, generates a draft with model-specific experience guidance, and applies dual consistency assurance. During verification, \textbf{\textsc{ConWriter}} internally constructs a provisional candidate state from the current scene without committing it to dynamic memory. 
If blocking violations are detected, \textbf{\textsc{ConWriter}} performs bounded sentence-level repair, with uncertainty-aware risk signals used to prioritize high-risk segments during repair.

\input{algos/ConWriter}

We omit detailed prompt templates because they contain implementation-specific control instructions. The algorithm, memory structure, transition operators, validation criteria, and repair procedure are described in the main paper and Algorithm~\ref{alg:conwriter}.

\section{Additional Experiment}
\label{app:Additional-Experiment}

\subsection{Dataset}
\label{app:Experiment_Dataset}

We evaluate \textbf{\textsc{ConWriter}} on ConStory-Bench~\cite{2026_arXiv_ConStory-Bench-dataset_Lost-in-Stories--Consistency-Bugs-in-Long-Story-Generation-by-LLMs}, a long-form story generation benchmark designed to evaluate narrative consistency. ConStory-Bench contains 2,000 prompts and covers four task scenarios:

\begin{itemize}
    \item \textbf{Generation}: writing a complete story from a minimal plot setup.
    \item \textbf{Continuation}: extending a given story fragment while preserving established facts and states.
    \item \textbf{Expansion}: expanding a concise plot outline into a detailed long-form story.
    \item \textbf{Completion}: filling the missing middle part between predefined beginning and ending constraints.
\end{itemize}

In the original ConStory-Bench setting, prompts are designed to elicit stories of approximately 8,000--10,000 words. In our experiments, we instead adopt a controlled forced-length setting with three target lengths: 3K, 6K, and 12K words. This design allows us to evaluate how narrative consistency changes as the required output length increases. The 3K setting tests short-to-medium-range consistency, the 6K setting evaluates mid-range narrative tracking, and the 12K setting stresses long-range dependencies over extended story development.

For each target length, all compared methods are given the same prompts and the same length requirement. This ensures that consistency differences are not caused by different output-length instructions. Since longer stories naturally introduce more entities, events, temporal relations, and stylistic commitments, the multi-length setting provides a more controlled way to analyze length-dependent consistency errors.

Due to the high cost of long-story generation and automatic consistency checking, we adopt a controlled evaluation subset. Specifically, we select the first five cases from each task type, resulting in 20 evaluation cases in total. This scale is consistent with prior long-form story generation work such as DOME~\cite{2025_NAACL_DOME_Generating-Long-form-Story-using-Dynamic-Hierarchical-Outlining-with-Memory-Enhancement}, which also evaluates on 20 generated long stories. All compared methods use the same prompts, task types, and length requirements. We do not modify the original ConStory-Bench prompts or task definitions.

\subsection{Target-Length Setting}
\label{sec:Appendix_Experiment_Setup_Target-Length}

We evaluate each method under three target lengths: 3K, 6K, and 12K words. For each setting, the generated story is required to reach the corresponding target length rather than stopping at the model's naturally preferred length. This differs from the original reporting setting in ConStory-Bench, where prompts are designed to elicit long stories but are not evaluated under the same controlled forced-length protocol.

Our forced-length setting makes the task stricter. As the required length increases, a model must introduce more events, entities, temporal relations, and stylistic commitments while preserving previously established facts and narrative states. This creates more opportunities for memory drift, transition inconsistency, and long-range factual conflicts. Therefore, the overall CED values in our experiments may be higher than those reported under less length-constrained generation settings.

This design allows us to compare methods under the same output-length requirement and analyze how consistency changes as the target length increases. Since CED is length-normalized, we also report average output words together with CED to avoid overestimating methods that generate shorter or simpler stories.

\begin{figure*}[th]
    \centering
    \includegraphics[width=0.26\linewidth]{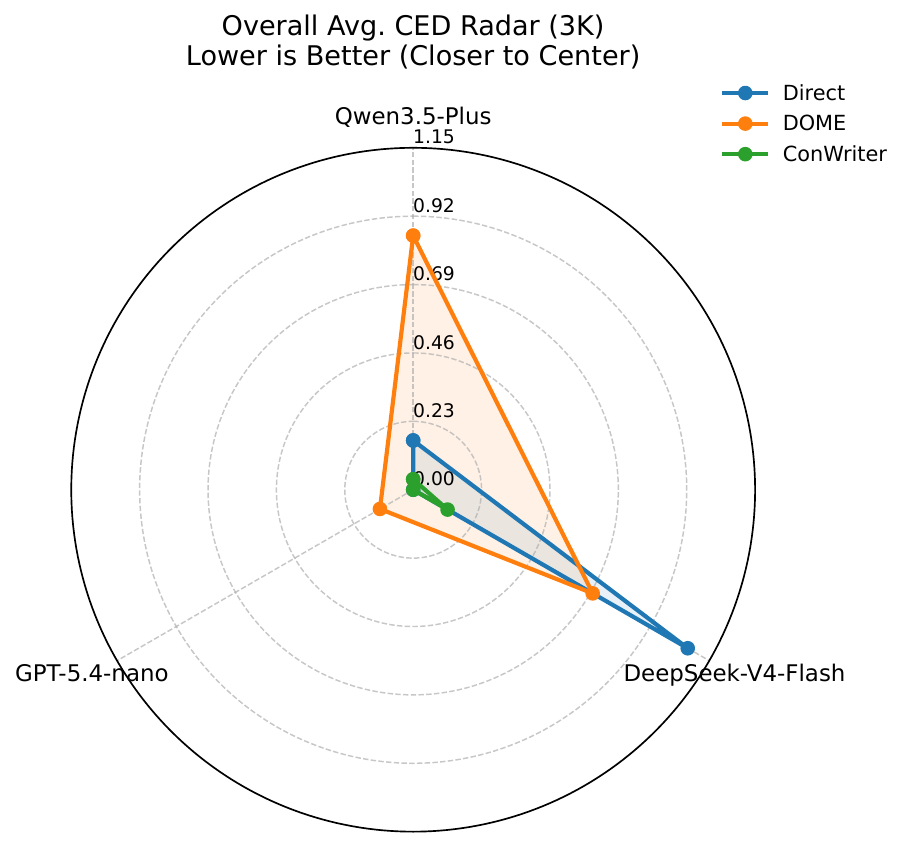}
    \includegraphics[width=0.26\linewidth]{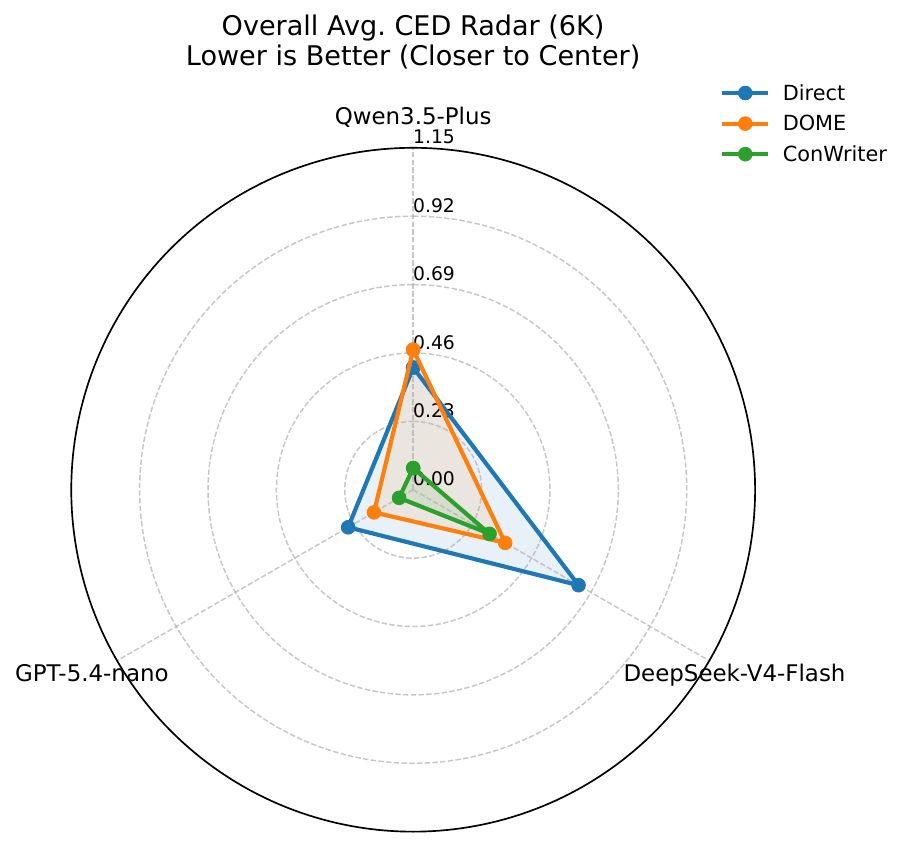}
    \includegraphics[width=0.26\linewidth]{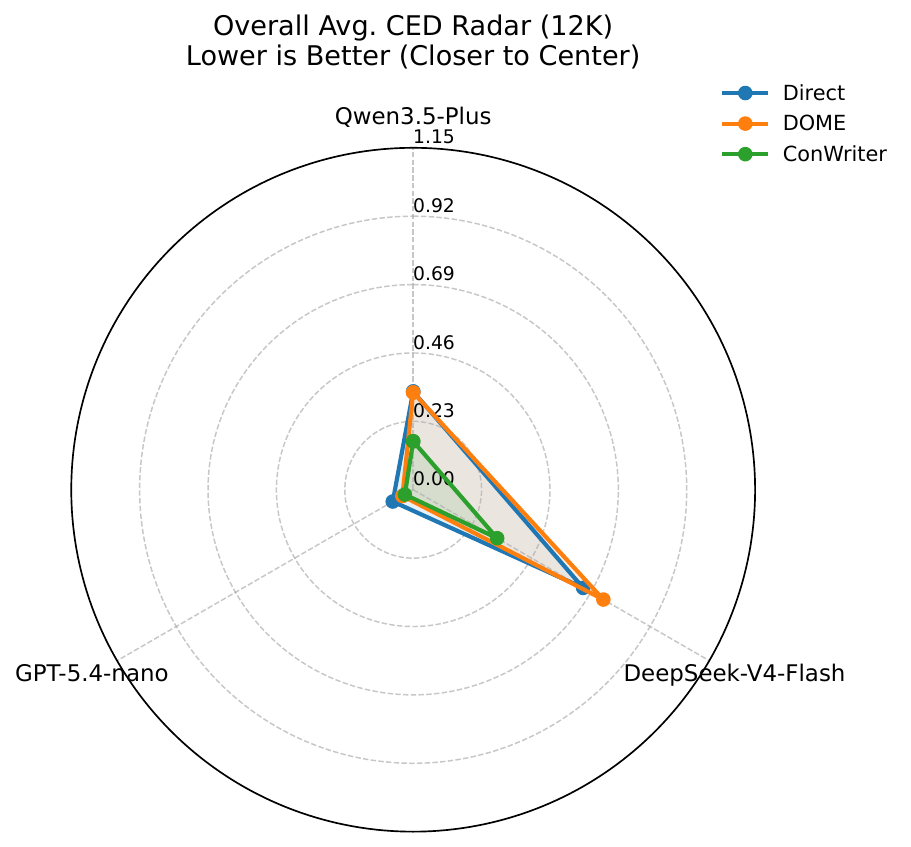}
    \caption{Length-wise radar visualization of Overall Avg. CED under the forced-length setting. Each axis corresponds to one target length, and lower values closer to the center indicate better consistency.}
    \label{fig:radar-3_Direct-DOME-ConWriter}
\end{figure*}

\input{tabs/Overall-avg-CED}

\subsection{Evaluation}
\label{app:Experiment_Evaluation}

We follow the official ConStory-Bench \cite{2026_arXiv_ConStory-Bench-dataset_Lost-in-Stories--Consistency-Bugs-in-Long-Story-Generation-by-LLMs} evaluation protocol and use ConStory-Checker to detect consistency errors in generated stories. ConStory-Checker is an evidence-grounded LLM-as-a-judge pipeline. It first extracts contradiction-prone spans, then pairs potentially conflicting spans, and finally outputs structured error reports with textual evidence and explanations.

Following ConStory-Bench, we use o4-mini~\cite{2025_OpenAI-System-Card_o4-mini_OpenAI-o3-and-o4-mini-System-Card} as the judge model. The evaluation covers five consistency facets:

\begin{itemize}
    \item \textbf{Timeline}: temporal order, duration, simultaneity, and causal progression.
    \item \textbf{Characterization}: character memory, knowledge, skills, personality, and abilities.
    \item \textbf{Basic Facts}: names, appearances, quantities, locations, objects, and factual details.
    \item \textbf{Commonsense}: physical plausibility, world rules, social norms, and causal commonsense.
    \item \textbf{Style}: narrative perspective, tone, and stylistic consistency.
\end{itemize}

We report the overall consistency error density and facet-level error densities. Let $e_i$ denote the number of detected consistency errors in story $i$, and $w_i$ denote the number of words in the generated story. The consistency error density is computed as:
\begin{equation}
\mathrm{CED}_i = \frac{e_i}{w_i / 10000}.
\end{equation}

The final score is averaged over all evaluated stories:
\begin{equation}
\mathrm{CED} = \frac{1}{N}\sum_{i=1}^{N}\mathrm{CED}_i,
\end{equation}
where $N$ is the number of evaluation cases. Lower CED indicates better narrative consistency.

Length normalization is necessary because longer stories naturally provide more opportunities for contradictions. Therefore, when a model generates a much shorter story than requested, we interpret its CED together with output length to avoid overestimating consistency due to under-generation.

\subsection{Additional Main Results Analysis}
\label{sec:Appendix_Experiment_Main-Results}

Table~\ref{tab:main_results} reports the main results on ConStory-Bench under the forced-length setting. Overall, \textbf{\textsc{ConWriter}} consistently reduces consistency error density across the three evaluated base LLM families, especially when the base model exhibits non-trivial consistency errors under direct generation. This shows that generation-time state tracking, symbolic transition checking, and localized repair can improve long-form story consistency beyond simply prompting the same model to generate a longer story.

For Qwen3.5-Plus, \textbf{\textsc{ConWriter}} improves the overall Avg. CED from 0.1657 to 0.0352 at 3K, from 0.4109 to 0.0728 at 6K, and from 0.3303 to 0.1629 at 12K. These results correspond to relative reductions of 78.76\%, 82.28\%, and 50.68\%, respectively. The improvement is particularly clear at 6K, where direct generation produces higher consistency errors in expansion and completion, while \textbf{\textsc{ConWriter}} substantially reduces both. Compared with DOME, \textbf{\textsc{ConWriter}} is also more stable: DOME increases the overall Avg. CED at 3K and 6K, while \textbf{\textsc{ConWriter}} consistently reduces it.

For DeepSeek-V4-Flash, \textbf{\textsc{ConWriter}} obtains larger gains. The overall Avg. CED is reduced from 1.0662 to 0.1335 at 3K, from 0.6421 to 0.2968 at 6K, and from 0.6602 to 0.3261 at 12K. These correspond to relative reductions of 87.48\%, 53.78\%, and 50.61\%. This result is important because DeepSeek-V4-Flash produces relatively long outputs under direct generation, but longer outputs also expose more opportunities for timeline, factual, and transition-level inconsistencies. \textbf{\textsc{ConWriter}} mitigates this issue by validating each scene before committing it into dynamic memory, rather than allowing local errors to propagate into later scenes.

For GPT-5.4-nano, direct generation is already very strong, especially at 3K where its Avg. CED is 0.0. This creates a ceiling-effect setting in which further improvement is difficult to observe. Even under this condition, \textbf{\textsc{ConWriter}} remains competitive: it preserves 0.0 Avg. CED at 3K, reduces Avg. CED from 0.2525 to 0.0546 at 6K, and reduces Avg. CED from 0.0793 to 0.0324 at 12K. These results suggest that \textbf{\textsc{ConWriter}} is most beneficial when the base model has observable consistency errors, while still avoiding degradation when the base model is already highly consistent.

The results also reveal the importance of considering output length together with CED. Under the forced-length setting, all methods are required to satisfy the specified target length, which makes consistency control more difficult than unconstrained generation. In several cases, \textbf{\textsc{ConWriter}} generates longer stories than direct generation while still reducing Avg. CED. For example, with Qwen3.5-Plus at 6K, \textbf{\textsc{ConWriter}} increases the average output length from 6721.05 to 7412.25 words and reduces Avg. CED from 0.4109 to 0.0728. With GPT-5.4-nano at 12K, it increases the average output length from 12492.35 to 16515.85 words and still reduces Avg. CED from 0.0793 to 0.0324. This indicates that the improvement is not merely caused by generating shorter or simpler stories. Instead, the framework can maintain lower consistency error density while supporting extended story development.

Across all settings, the strongest improvements appear when direct generation suffers from accumulated state drift or transition inconsistency. These gains support the central hypothesis of this work: long-story consistency requires explicit generation-time control over memory, state transitions, and local repair, rather than relying only on a stronger base LLM or a longer context window.

\subsection{Additional Ablation Study Analysis}
\label{sec:Appendix_Experiment_Ablation-Study}

Table~\ref{tab:ablation} reports the ablation results under the 3K forced-length setting with DeepSeek-V4-Flash. We choose this setting because DeepSeek-V4-Flash exhibits clear consistency errors under direct generation, making it suitable for analyzing the contribution of each component.

Removing dynamic memory leads to a substantial degradation. The overall Avg. CED increases from 0.1335 to 0.7499. The degradation is especially clear on the generation and expansion tasks, where Avg. CED increases from 0.0 to 0.9135 and from 0.0 to 1.6309, respectively. This confirms that dynamic memory is essential for tracking evolving events, entity states, relations, and unresolved constraints. Without dynamic memory, the system cannot reliably determine whether a newly generated scene is compatible with the current narrative state, and later scenes are more likely to introduce contradictions.

Removing structured validation also causes a large performance drop, increasing the overall Avg. CED from 0.1335 to 0.6843. This variant still has access to memory, but it lacks explicit transition checking and violation anchoring. As a result, the model may retrieve relevant story information but fail to verify whether the generated scene actually satisfies required preconditions, postconditions, and forbidden constraints. The strongest degradation appears in the completion task, where Avg. CED increases from 0.0 to 1.4995, suggesting that structured validation is particularly important when the generated content must bridge strict beginning and ending constraints.

Removing uncertainty monitoring increases the overall Avg. CED from 0.1335 to 0.5463. This shows that uncertainty-aware risk signals provide useful complementary information beyond symbolic checks. Although the uncertainty monitor is not treated as a hard consistency judge, it helps identify unstable or weakly grounded sentences that deserve stronger validation or localized repair. Without this signal, some risky segments may pass through the generation loop without sufficient inspection, leading to more residual errors.

The full \textbf{\textsc{ConWriter}} achieves the best overall result among all variants, with an Avg. CED of 0.1335. The ablation results show that the three components play complementary roles. Dynamic memory provides stateful narrative context, structured validation turns scene generation into a checkable transition process, and uncertainty monitoring prioritizes high-risk regions for additional verification and repair. Removing any of them weakens the overall consistency-control loop.

\input{tabs/cost-analysis}

\subsection{Computational Cost Analysis}
\label{app:cost_analysis}

We further analyze the inference cost of \textsc{ConWriter} using the first five cases of Task 1 (Continuation). Table~\ref{tab:cost_analysis} reports the average output length, number of API calls, and token consumption per generated story. Compared with Direct generation and DOME, \textsc{ConWriter} introduces additional inference overhead because its scene-level generation pipeline performs planning, consistency checking, and localized repair. The magnitude of this overhead varies across base LLMs, reflecting differences in instruction following, revision behavior, and the number of required generation and repair steps. We therefore view the additional computational cost as a trade-off for generation-time consistency control.

\subsection{Additional Discussion}
\label{sec:Appendix_Experiment_Discussion}

\subsubsection{Planning and memory can help, but only when the control pipeline is reliable.}
Our results show that planning and memory mechanisms are useful for base LLMs whose direct generation is less stable. For such models, external scaffolding can provide explicit structure, remind the model of previous story states, and reduce uncontrolled narrative drift. However, this effect is not guaranteed for stronger LLMs. When a model already has strong implicit planning, memory tracking, and reasoning ability, a weakly constrained planning or memory pipeline may become a source of interference rather than assistance. For example, DOME sometimes improves direct generation, but it can also increase CED under several settings. This suggests that simply adding a plan or memory is insufficient; the external pipeline must also verify whether generated scenes actually satisfy evolving narrative states. \textbf{\textsc{ConWriter}} addresses this issue by combining dynamic memory with symbolic transition validation and localized repair, making the additional control more targeted and less disruptive.

\subsubsection{Why evaluate on strong contemporary LLMs?}
We focus on relatively strong base LLMs because long-form story generation requires broad language modeling, narrative planning, commonsense reasoning, stylistic control, and long-range coherence. Very weak models are unlikely to generate high-quality long stories, so their errors may reflect general generation failure rather than the specific consistency problem studied in this paper. Meanwhile, training or fine-tuning a weaker model for long-story generation would require substantial data, computation, and engineering cost. In contrast, a training-free LLM-agent framework is more practical: it treats the base LLM as a capable generator and improves reliability through generation-time memory, validation, and repair. This setting also better reflects realistic usage, where strong closed-source or API-based models are commonly used for long-form writing but still suffer from consistency drift.

\subsubsection{Length usually increases risk, but CED is not strictly monotonic.}
A natural expectation is that longer stories should produce higher consistency error density because they introduce more events, entities, temporal relations, and factual commitments. This tendency appears in several settings, especially when the base model accumulates state drift over longer generations. However, the trend is not strictly monotonic for every model. Some models obtain lower CED at 12K than at 3K. This does not mean longer generation is easier; rather, CED is affected by model-specific generation behavior, task difficulty, local story structure, and the limited controlled subset. Since CED is length-normalized, a longer story can also dilute a small number of detected errors if the generation remains coherent. Therefore, we interpret length as a general risk factor rather than a deterministic predictor of CED. The main finding remains that \textbf{\textsc{ConWriter}} reduces CED more consistently than Direct and DOME across model families and target lengths.

\subsubsection{Repair compliance under strong LLMs.}
We also observe an interesting failure mode for GPT-5 and GPT-5-mini. Although these models have strong planning, memory, and reasoning abilities under direct generation, they are less stable when repeatedly corrected by the dual-checking module in longer \textsc{ConWriter} runs. When many violations are reported, the model sometimes fails to revise the specified errors, or revises them at the cost of violating the required target length. This suggests a conflict between local corrective feedback and global length control: repeated repair instructions can make a strong model over-focus on satisfying the verifier, while weakening its adherence to the long-form generation requirement. As a result, GPT-5 \textsc{ConWriter} is feasible in the 3K setting, but does not reliably converge under the 6K and 12K settings. We therefore report only the feasible GPT-5 \textsc{ConWriter} result in Table~\ref{tab:main_results}.

\section{LLM Usage Statement}
\label{sec:Appendix_LLM-Usage-Statement}

We use LLM-based tools to assist with language polishing, grammar checking, and minor presentation support, such as refining figure captions and helping design small visual elements for figures. We develop, select, and verify the research idea, problem formulation, method design, experimental setup, analysis, and final scientific claims. LLM tools do not autonomously generate the core research contributions or determine experimental conclusions. OpenAI Codex~\cite{openai2025codex} is additionally used for limited implementation assistance and code-level inspection, and we review and validate all resulting code changes and experimental pipelines.

%% file: sec-2_Related-Work.tex
\section{Related Work}
\label{sec:Related-Work}

\subsection{Long-Form Story Generation}

Long-form story generation requires models to maintain coherent plots, stable characters, consistent timelines, and long-range narrative dependencies across extended contexts. Existing studies have explored this problem from several perspectives, including outline control, memory-enhanced generation, cognitive writing, reflection-based revision, and training-based long-text generation. 
For instance, 
DOC~\cite{2023_ACL_DOC_DOC--Improving-Long-Story-Coherence-with-Detailed-Outline-Control} improves long-story coherence by decomposing story generation into detailed outline construction and outline-guided writing. 
LongStory~\cite{2024_PAKDD_LongStory_LongStory--Coherent-Complete-and-Length-Controlled-Long-Story-Generation} combines context-weight calibration with structural position modeling, improving length-controlled story generation.
More recent work introduces more adaptive planning and memory mechanisms. DOME~\cite{2025_NAACL_DOME_Generating-Long-form-Story-using-Dynamic-Hierarchical-Outlining-with-Memory-Enhancement} generates stories with dynamic hierarchical outlines and memory enhancement, allowing the model to organize long-form content at multiple levels of abstraction. CogWriter~\cite{2025_ACL-Findings_CogWriter_A-Cognitive-Writing-Perspective-for-Constrained-Long-Form-Text-Generation} studies constrained long-form generation from a cognitive writing perspective, emphasizing the role of structured constraints in guiding extended writing. 
SuperWriter~\cite{2025_arXiv_SuperWriter_SuperWriter--Reflection-Driven-Long-Form-Generation-with-Large-Language-Models} explores reflection-driven generation, where the model iteratively evaluates and revises its own outputs. 
Storywriter~\cite{2025_CIKM_StoryWriter_StoryWriter--A-Multi-Agent-Framework-for-Long-Story-Generation} is a collaborative multi-agent framework, designing outlining, planning, and writing agents for long story generation.
LongWriter-Zero~\cite{2025_ICLR_LongWriter-Zero_LongWriter-Zero--Mastering-Ultra-long-Text-Generation-via-Reinforcement-Learning} takes a different direction by improving ultra-long text generation through reinforcement learning.

These methods demonstrate the importance of planning, memory, reflection, and training for long-form generation. However, most of them focus on improving global coherence or output length, rather than explicitly verifying whether each generated scene satisfies evolving narrative states. In contrast, \textbf{\textsc{ConWriter}} focuses on training-free generation-time consistency control. It writes stories incrementally at the scene level, maintains structured narrative memory, performs symbolic state-transition checking, uses uncertainty-aware risk monitoring, and applies local repair before committing new content into dynamic memory.

\subsection{LLM/Agent Memory}

Memory has become a central component of LLM agents, especially for long-horizon interaction, personalization, task continuity, and context-aware reasoning~\cite{2026_ACL_HeLa-Mem_HeLA-Mem--Hebbian-Learning-and-Associative-Memory-for-LLM-Agents}. Recent surveys summarize memory mechanisms from both human-inspired and agent-system perspectives. The human-memory perspective~\cite{2025_arXiv_Survey_From-Human-Memory-to-AI-Memory--A-Survey-on-Memory-Mechanisms-in-the-Era-of-LLMs} relates AI memory to cognitive notions such as short-term, long-term, episodic, semantic, and procedural memory. The agent-system perspective~\cite{2025_arXiv_Survey_Memory-in-the-Age-of-AI-Agents} further discusses how memory supports autonomous agents through storage, retrieval, update, reflection, and long-term adaptation.

Representative memory systems instantiate these ideas in different ways. 
For instance, MemoryBank~\cite{2024_AAAI_MemoryBank_MemoryBank--Enhancing-Large-Language-Model-with-Long-term-Memory} equips LLMs with long-term memory to preserve user-related information across interactions, supporting more personalized and consistent responses. Mem0~\cite{2025_arXiv_Mem0_Mem0--Building-Production-Ready-AI-Agents-with-Scalable-Long-Term-Memory} provides a scalable memory framework for production-ready AI agents, emphasizing efficient storage, retrieval, and update of relevant memories. 
A-MEM \cite{2025_NeurlPS_A-Mem_A-Mem-Agentic-Memory-for-LLM-Agents} combines Zettelkasten-style knowledge linking with agent-driven memory update mechanisms for long-term memory.
CA3Mem~\cite{2026_AAAI_CA3Mem_Evolving-Generalist-Virtual-Agents-with-Generative-and-Associative-Memory} transforms agent memory from passive experience storage into a generative graph-based mechanism, enabling associative and generative memory.

These works show that memory is essential for extending LLMs beyond single-turn generation. However, general agent memory systems are often designed for personalization, retrieval augmentation, or knowledge organization. \textbf{\textsc{ConWriter}} uses memory for a more specific purpose: narrative consistency control in long story generation. Static memory preserves global story commitments such as premise, character profiles, world rules, and style constraints, while dynamic memory records evolving events, entity states, relations, timelines, and unresolved constraints. This dual-memory design enables \textbf{\textsc{ConWriter}} to verify scene transitions and repair local inconsistencies before they propagate into later parts of the story.

%% file: algos/ConWriter.tex
\begin{algorithm}[t]
\caption{Procedure of \textbf{\textsc{ConWriter}}}
\label{alg:conwriter}
\scriptsize
\begin{algorithmic}[1]

\Require Story input $x$, base LLM $G_{\theta}$,
experience bank $\mathcal{E}_{\theta}$
\Ensure Long-Form story output $Y$

\State Construct scene specifications $\{s_1,\ldots,s_T\}$ from $x$
\State Initialize static memory $\mathcal{M}^{s}$,
dynamic memory $\mathcal{M}^{d}_{1}$, and $Y \leftarrow [\ ]$

\For{$t=1$ to $T$}

    \State $\mathcal{M}_t \leftarrow
    (\mathcal{M}^{s},\mathcal{M}^{d}_{t})$

    \State $\mathcal{R}_t \leftarrow
    \operatorname{Retrieve}(s_t,\mathcal{M}_t)$

    \State $o_t \leftarrow
    \operatorname{Reason}(s_t,\mathcal{R}_t)$
    \Comment{derive symbolic transition constraints}

    \State Select scene-specific experience guidance
    $\mathcal{E}_t$ from $\mathcal{E}_{\theta}$

    \State $\hat{y}_t \leftarrow
    G_{\theta}(x,s_t,\mathcal{R}_t,o_t,\mathcal{E}_t)$

    \State $(v_t,r_t) \leftarrow
    \operatorname{Assure}
    (\hat{y}_t,\mathcal{M}_t,o_t)$

    \State $y_t \leftarrow
    \operatorname{AcceptOrRepair}
    (\hat{y}_t,v_t,r_t)$
    \Comment{bounded sentence-level repair if needed}

    \If{$y_t$ is accepted}

        \State $\Delta\mathcal{M}^{d}_{t} \leftarrow
        \operatorname{Extract}
        (y_t,\mathcal{M}^{s},\mathcal{M}^{d}_{t},o_t)$

        \If{$\operatorname{Commit}
        (\Delta\mathcal{M}^{d}_{t})=1$}

            \State $\mathcal{M}^{d}_{t+1} \leftarrow
            \mathcal{M}^{d}_{t}
            \oplus
            \Delta\mathcal{M}^{d}_{t}$

            \State $Y.\operatorname{append}(y_t)$

        \Else
            \State $\mathcal{M}^{d}_{t+1}
            \leftarrow \mathcal{M}^{d}_{t}$
        \EndIf

    \Else
        \State $\mathcal{M}^{d}_{t+1}
        \leftarrow \mathcal{M}^{d}_{t}$
    \EndIf

\EndFor

\State \Return $\operatorname{Concat}(Y)$

\end{algorithmic}
\end{algorithm}

%% file: tabs/Overall-avg-CED.tex
\begin{table*}[th]
\centering
\renewcommand{\arraystretch}{1.3}
\scalebox{0.66}{
\begin{tabular}{llcccc}
\toprule
Target Length & Method & Qwen3.5-Plus & DeepSeek-V4-Flash & GPT-5.4-nano & Avg. \\
\midrule
\multirow{3}{*}{3K}
& Direct     & 0.1657 & 1.0662 & 0.0000 & 0.4106 \\
& DOME       & 0.8546 & 0.6971 & 0.1291 & 0.5603 \\
& \textbf{\textsc{ConWriter}} & \textbf{0.0352} & \textbf{0.1335} & \textbf{0.0000} & \textbf{0.0562} \\
\midrule
\multirow{3}{*}{6K}
& Direct     & 0.4109 & 0.6421 & 0.2525 & 0.4352 \\
& DOME       & 0.4709 & 0.3572 & 0.1522 & 0.3268 \\
& \textbf{\textsc{ConWriter}} & \textbf{0.0728} & \textbf{0.2968} & \textbf{0.0546} & \textbf{0.1414} \\
\midrule
\multirow{3}{*}{12K}
& Direct     & 0.3303 & 0.6602 & 0.0793 & 0.3566 \\
& DOME       & 0.3272 & 0.7386 & 0.0419 & 0.3692 \\
& \textbf{\textsc{ConWriter}} & \textbf{0.1629} & \textbf{0.3261} & \textbf{0.0324} & \textbf{0.1738} \\
\bottomrule
\end{tabular}
}
\caption{Overall Avg. CED of Direct, DOME, and \textbf{\textsc{ConWriter}} across three base LLM families and target lengths. The Avg. column reports the mean over the three base LLM families. Lower values indicate better consistency.}
\label{tab:overall_avg_CED}
\end{table*}

%% file: tabs/cost-analysis.tex
\begin{table*}[t]
\centering
\small
\setlength{\tabcolsep}{8pt}
\begin{tabular}{llrrr}
\toprule
\textbf{Base LLM} & \textbf{Method} &
\textbf{Avg. words/story} &
\textbf{Avg. API calls/story} &
\textbf{Avg. tokens/story} \\
\midrule
\multirow[c]{3}{*}{Qwen3.5-Plus}
& Direct    & 16,086.0 & 27.0 & 375,833.0 \\
& DOME      & 12,583.6 & 31.6 & 181,758.6 \\
& \textsc{ConWriter} & 17,204.2 & 79.8 & 1,158,532.2 \\
\midrule
\multirow[c]{3}{*}{DeepSeek-V4-Flash}
& Direct    & 16,082.8 & 20.6 & 62,957.8 \\
& DOME      & 11,904.2 & 31.6 & 90,067.0 \\
& \textsc{ConWriter} & 14,778.0 & 53.2 & 495,749.6 \\
\midrule
\multirow[c]{3}{*}{GPT-5.4-nano}
& Direct    & 16,228.8 & 9.6  & 39,362.0 \\
& DOME      & 12,647.0 & 16.0 & 54,453.2 \\
& \textsc{ConWriter} & 19,103.0 & 32.8 & 508,538.6 \\
\bottomrule
\end{tabular}
\caption{Computational cost on the first five cases of Task 1 (Continuation). Values are averaged per generated story.}
\label{tab:cost_analysis}
\end{table*}